\documentclass[11pt]{article}

\usepackage{graphicx}
\usepackage{multirow}
\usepackage{amsmath,amssymb,amsfonts}
\usepackage{amsthm}
\usepackage{mathrsfs}
\usepackage[title]{appendix}
\usepackage{xcolor}
\usepackage{textcomp}
\usepackage{manyfoot}
\usepackage{booktabs}
\usepackage{algorithm}
\usepackage{algorithmicx}
\usepackage{algpseudocode}
\usepackage{listings}
\usepackage{braket}
\usepackage{tikz}
\usepackage{float}
\usepackage{placeins}
\usepackage[
    colorlinks=true,
    linkcolor=blue,
    citecolor=blue,
    urlcolor=blue
]{hyperref}
\usepackage{subcaption}

\newtheorem{remark}{Remark}%

\usepackage{authblk}

\title{Spectral Phase Encoding for Quantum Kernel Methods}

\author[1]{Pablo Herrero Gómez}
\author[1]{Antonio Jimeno Morenilla}
\author[1]{David Muñoz-Hernández}
\author[1]{Higinio Mora Mora}
\affil[1]{Department of Computer Science Technology and Computation, University of Alicante, Alicante, Spain}

\date{}

\begin{document}
\maketitle

\begin{abstract}
Quantum kernel methods are promising for near-term quantum machine learning, yet their behavior under data corruption remains insufficiently understood. We analyze how quantum feature constructions degrade under controlled additive noise.

We introduce \emph{Spectral Phase Encoding} (SPE), a hybrid construction combining a discrete Fourier transform (DFT) front-end with a diagonal phase-only embedding aligned with the geometry of diagonal quantum maps. Within a unified framework, we compare QK-DFT against alternative quantum variants (QK-PCA, QK-RP) and classical SVM baselines under identical clean-data hyperparameter selection, quantifying robustness via dataset fixed-effects regression with wild cluster bootstrap inference across heterogeneous real-world datasets.

Across the quantum family, DFT-based preprocessing yields the smallest degradation rate as noise increases, with statistically supported slope differences relative to PCA and RP. Compared to classical baselines, QK-DFT shows degradation comparable to linear SVM and more stable than RBF SVM under matched tuning. Hardware experiments confirm that SPE remains executable and numerically stable for overlap estimation. These results indicate that robustness in quantum kernels depends critically on structure-aligned preprocessing and its interaction with diagonal embeddings, supporting a robustness-first perspective for NISQ-era quantum machine learning.
\end{abstract}


\section{Introduction}

Quantum machine learning (QML) has attracted considerable attention as a potential
avenue for leveraging near-term quantum devices in supervised learning tasks.
Among the various paradigms explored, quantum kernel methods have emerged as a
particularly appealing approach due to their conceptual simplicity and natural
compatibility with noisy intermediate-scale quantum (NISQ) hardware.
By embedding classical data into quantum states and estimating inner products in
an implicit feature space, quantum kernels provide a principled bridge between
classical learning theory and quantum computation
\cite{schuld_is_2022,havlicek_supervised_2019}.

Despite this promise, it is now well understood that the practical performance of
quantum kernel methods is strongly influenced by the choice of data encoding.
Recent studies have shown that kernel behavior is often dominated not by the
quantum model itself, but by how classical data are mapped into the quantum feature
space \cite{rath_quantum_2024}.
Inappropriate encodings can lead to trivial kernels, excessive concentration, or
poor generalization, thereby obscuring any potential quantum advantage
\cite{thanasilp_exponential_2024}.
As a result, the design of informative and noise-resilient encodings has become a
central challenge for quantum kernel methods in realistic regimes.

At the same time, there is growing recognition that quantum advantage should not
be treated as a universal or binary objective.
Rather than focusing exclusively on outperforming classical baselines, recent
work emphasizes identifying regimes in which quantum models are practically
viable, interpretable, and robust under noise and finite sampling
\cite{schuld_is_2022,jerbi_quantum_2023}.
From this perspective, understanding how quantum pipelines degrade under
controlled perturbations is at least as important as reporting peak performance.

A recurring insight in this line of research is the importance of
\emph{structure-aware} quantum kernels.
Encodings that respect symmetries, invariances, or internal organization of the
data have been shown to mitigate kernel concentration and improve robustness
\cite{glick_covariant_2024}.
These observations suggest that incorporating inductive bias at the encoding stage
is a promising strategy for enhancing the practical utility of quantum kernels,
particularly in the NISQ regime.

In classical machine learning, one of the most successful ways to introduce such
inductive bias is through preprocessing.
Techniques such as principal component analysis (PCA) or frequency-domain
representations based on the discrete Fourier transform (DFT) are routinely used
to denoise data, reduce dimensionality, and expose global structure.
In particular, spectral representations have proven especially effective for
structured data such as time series, signals, and images, where meaningful
patterns often manifest more clearly in the frequency domain.
While these ideas are well established classically, their systematic integration
into quantum kernel pipelines remains comparatively underexplored.

Motivated by these observations, this work focuses on the role of spectral
structure in quantum kernel methods.
We introduce \emph{Spectral Phase Encoding} (SPE), a frequency-aware encoding that
combines a discrete Fourier transform with diagonal quantum gates to encode
spectral coefficients as phases.
This choice leverages the natural expressivity of diagonal unitaries while
preserving the global structure revealed by spectral preprocessing.

Rather than aiming to maximize accuracy at a fixed noise level, we adopt a
noise-aware evaluation strategy that explicitly characterizes how performance
degrades as noise increases.
Through a systematic experimental study on synthetic benchmarks, real-world
datasets, and real quantum hardware, we analyze when spectral phase encoding is
beneficial, when its advantages vanish, and how it behaves under realistic device
noise.
This approach allows us to delineate the scope of applicability and limitations of
SPE in a principled and empirically grounded manner.

\paragraph*{Related Work}
Quantum kernel methods were initially formalized as learning algorithms operating
in implicit quantum feature spaces, enabling nonlinear classification through inner
products estimated on quantum devices
\cite{schuld_is_2022,havlicek_supervised_2019}.
This formulation established a direct connection between classical kernel theory
and quantum computation, and motivated a broad line of research on the design,
evaluation, and deployment of quantum kernels under realistic conditions.

Subsequent work has investigated the feasibility of training and deploying quantum
kernels on near-term hardware.
In particular, Hubregtsen et al.\ demonstrated that quantum embedding kernels can be
trained on noisy devices, while highlighting the strong sensitivity of kernel
performance to hardware noise, finite sampling, and circuit depth
\cite{hubregtsen_training_2022}.
Related studies have emphasized that practical performance is often limited not by
the expressivity of the quantum model itself, but by the interaction between the
chosen encoding, noise processes, and measurement statistics.

A parallel line of work has focused on improving the efficiency and scalability of
quantum kernel methods.
Sahin et al.\ proposed sub-sampling strategies to reduce the computational cost of
kernel alignment and training, enabling larger-scale experiments within realistic
resource budgets \cite{sahin_efficient_2024}.
Variational and hybrid approaches, such as approximate or trainable QSVM
formulations, have also been explored as a means of balancing expressivity and
resource requirements in the NISQ regime \cite{park_variational_2023}.

Beyond discrete-variable encodings, continuous-variable quantum kernels have been
investigated as an alternative approach to representing classical data, offering
greater flexibility at the cost of increased experimental complexity
\cite{henderson_quantum_2024}.
Across both discrete- and continuous-variable settings, a recurring limitation is
the phenomenon of kernel concentration, whereby quantum kernels become nearly
constant and lose discriminative power as system size, data dimension, or noise
levels increase \cite{thanasilp_exponential_2024}.
This observation has motivated the development of structure-aware and
symmetry-respecting kernels, which incorporate inductive bias at the encoding stage
to mitigate concentration effects and improve robustness
\cite{glick_covariant_2024}.

The importance of data encoding has been further underscored by theoretical studies
analyzing the expressive power and inductive bias of quantum feature maps.
It has been shown that the choice of encoding can fundamentally determine whether a
quantum kernel exhibits meaningful variability or collapses to a trivial regime,
independently of the downstream learning algorithm
\cite{rath_quantum_2024,schuld_is_2022}.
These results highlight encoding design as a central bottleneck for practical
quantum kernel methods.

In classical machine learning, the introduction of inductive bias is often achieved
through preprocessing.
Techniques such as principal component analysis (PCA) and frequency-domain
representations based on the discrete Fourier transform (DFT) are routinely used to
reduce dimensionality, suppress noise, and expose global structure in data.
Spectral representations, in particular, play a central role in time series
analysis, signal processing, and image modeling, where relevant information often
manifests more clearly in the frequency domain.
Recent advances in classical learning have further reinforced the effectiveness of
frequency-aware representations for structured data.

Despite their widespread use in classical pipelines, such preprocessing strategies
have received comparatively limited attention in the context of quantum kernel
methods.
Most existing quantum encodings operate directly on raw or minimally processed
features, without explicitly exploiting spectral structure or other domain-specific
regularities.
As a result, the potential benefits of combining classical structure-extracting
transformations with quantum feature maps remain insufficiently characterized.

The present work builds on these insights by investigating spectral structure as a
concrete and widely applicable source of inductive bias for quantum kernel
encodings.
Rather than proposing a new kernel learning paradigm, we focus on systematically
characterizing how frequency-aware phase encodings behave under increasing noise
and realistic hardware constraints, and on identifying the regimes in which such
structure-aware encodings remain informative.

\section{Methods}
\subsection{Spectral Phase Encoding}
\label{subsec:methods_spe}

We introduce \emph{Spectral Phase Encoding} (SPE), a structured quantum kernel construction that combines a discrete Fourier transform (DFT) front-end with a diagonal-gate-based phase embedding. SPE is designed to exploit a structural synergy between spectral representations and phase-only quantum encodings. Throughout the experimental section, this construction is denoted as \texttt{QK-DFT} to emphasize the separation between classical front-end and quantum backend within the unified regression framework used for robustness analysis.

The design of SPE is motivated by two complementary observations.
First, many real-world datasets exhibit meaningful structure in the frequency
domain, which can be revealed through classical spectral transformations.
Second, diagonal quantum gates provide a natural and hardware-efficient mechanism
for encoding relative phase information, which plays a central role in quantum
interference-based representations.

Classical machine learning has long exploited frequency-domain representations as
a form of preprocessing or feature extraction, particularly in signal processing,
time-series analysis, and image-based tasks.
DFT and related spectral representations are commonly
used to concentrate structured information into a reduced number of components,
highlighting periodicities, global correlations, and scale-dependent patterns
\cite{fulcher2014hctsa, yi2025surveyft}.
Recent work in deep learning for time series has further emphasized the role of
Fourier front-ends as a means of exposing inductive biases related to
shift-invariance and long-range dependencies \cite{zhou2022fedformer}.

Given an input vector $x \in \mathbb{R}^d$, SPE begins by applying a DFT, yielding a set of complex spectral coefficients that
encode global correlations and oscillatory patterns present in the data.
Rather than treating the DFT as part of a trainable model, as in frequency-aware
neural architectures, here it is used explicitly as a fixed preprocessing step
that defines the representation supplied to the quantum embedding.
This design mirrors classical pipelines in which spectral features are computed
prior to classification or kernel evaluation \cite{fulcher2014hctsa}.

Let $F(x) \in \mathbb{C}^d$ denote the discrete Fourier transform of $x$.
We retain the first $m$ spectral components and construct a phase vector
$\boldsymbol{\phi} \in [-\pi,\pi)^m$, obtained from the arguments of the complex
coefficients, optionally modulated by a bounded function of their magnitudes.
These phases define the diagonal unitary in Eq.~\eqref{eq:diag_unitary},
which is applied to a uniform superposition state prepared by Hadamard gates.
\begin{equation}
\label{eq:diag_unitary}
D(\boldsymbol{\phi}) =
\mathrm{diag}\!\left(e^{i\phi_0}, e^{i\phi_1}, \dots, e^{i\phi_{m-1}}\right),
\end{equation}
which is applied to a uniform superposition state prepared by Hadamard gates.
Diagonal unitaries admit particularly efficient decompositions, often avoiding
ancilla overhead and enabling low-depth implementations \cite{welch_diagonal_unitaries_2014}.
When $m$ is not a power of two, the phase vector is deterministically padded,
allowing the encoding to be implemented using
$n = \lceil \log_2 m \rceil$ qubits.

The resulting quantum state takes the form:
\begin{equation}
\ket{\psi(x)} =
\frac{1}{\sqrt{2^n}}
\sum_{j=0}^{2^n-1} e^{i \phi_j(x)} \ket{j},
\end{equation}
where $\phi_j(x)$ denotes either a retained or padded phase component.
This construction highlights a key advantage of diagonal-gate-based embeddings:
$2^n$ effective features are represented using only $n$ qubits, without requiring
state preparation procedures whose depth scales with the data dimension.
In contrast to amplitude encoding, which requires normalization and multi-qubit
controlled operations, SPE relies exclusively on single-qubit Hadamards and a
single diagonal unitary, resulting in shallow circuits that are well aligned with
current NISQ hardware constraints \cite{havlicek_supervised_2019}.
Moreover, in superconducting platforms, $Z$-axis phase rotations are commonly
implemented as ``virtual-$Z$'' updates with effectively zero duration and high
fidelity, making phase/diagonal constructions especially hardware-friendly
\cite{mckay_efficient_z_2017,kjaergaard_superconducting_2020,vezvaee_virtualz_2025}.

From a mathematical perspective, the combination of a Fourier front-end with
phase-only encoding is particularly natural.
The Fourier transform diagonalizes convolutional and shift-invariant operators,
concentrating structured information into a reduced number of spectral modes.
Encoding this information as relative phases preserves interference patterns while
remaining invariant under global amplitude rescaling.
Physically, diagonal gates implement relative phase accumulation with reduced
reliance on entangling operations, which mitigates circuit depth and exposure to
multi-qubit noise on NISQ devices \cite{mckay_efficient_z_2017,kjaergaard_superconducting_2020}.

In this work, SPE refers specifically to the combination of a DFT front-end with a diagonal-gate-based phase embedding. To isolate the contribution of the spectral structure, we also evaluate alternative front-ends—principal component analysis (PCA) and random projections (RP)—under the same diagonal quantum embedding, leading to the constructions QK-PCA and QK-RP in the experimental section.  This controlled comparison enables us to disentangle the effect of spectral preprocessing from that of the quantum embedding itself.

\begin{figure}[t]
    \centering
    \includegraphics[width=1\linewidth]{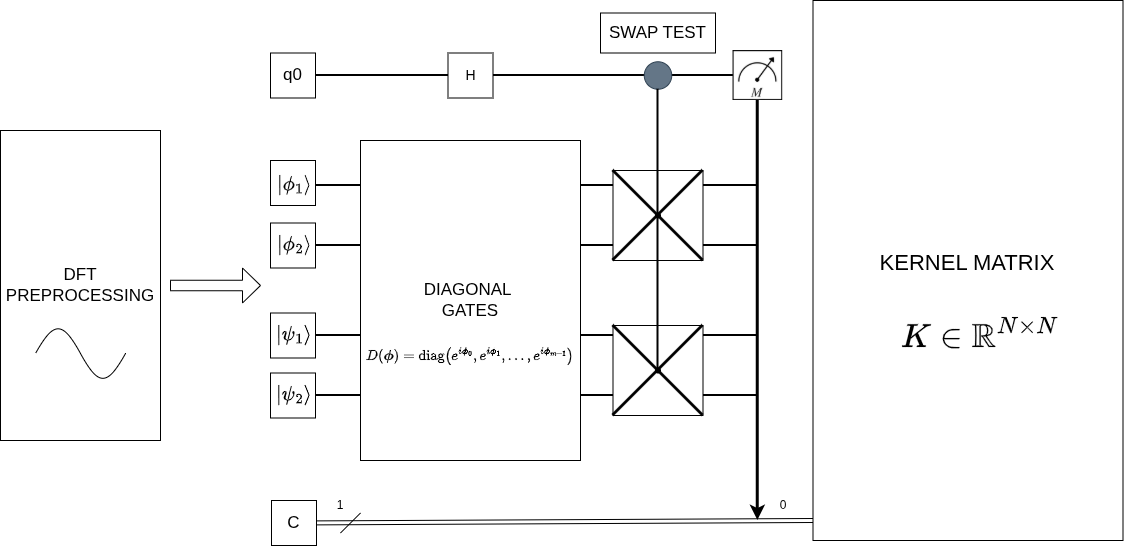}    
    \caption{Conceptual pipeline of Spectral Phase Encoding. Classical data are transformed to the frequency domain, mapped to a vector of relative phases, and embedded into a quantum state using a diagonal gate acting on a uniform superposition. Kernel values are obtained via overlap estimation.
    }
    \label{fig:spe-pipeline}
\end{figure}

Finally, kernel values are obtained by estimating overlaps between SPE-encoded
quantum states using a SWAP test.
This choice ensures direct compatibility with quantum kernel methods while
maintaining a clear separation between classical preprocessing, quantum state
preparation, and kernel evaluation. A schematic overview of the full SPE pipeline is shown in Fig.~\ref{fig:spe-pipeline}.
\begin{remark}[Phase-only encodings and complex-valued features]
\label{rem:complex_phase_encoding}
A practical motivation for diagonal-gate-based phase encoding is its natural
compatibility with complex-valued features. Spectral representations produced by the discrete Fourier transform are inherently complex, with relevant information encoded in both magnitude and phase. Most standard quantum feature maps, including amplitude and angle encodings, are defined for real-valued inputs and therefore require ad hoc transformations or the explicit discarding of complex phase information \cite{schuld_is_2022,havlicek_supervised_2019}. 
By contrast, diagonal quantum gates natively implement relative phase shifts,
providing a direct and hardware-efficient mechanism to encode complex phases.
This makes phase-only encodings particularly well suited for spectral
representations and allows the Fourier front-end in SPE to be integrated without
additional approximations, while preserving a low-depth, NISQ-compatible circuit
structure.
\end{remark}

\subsection{Kernel evaluation and learning protocol}
\label{subsec:methods_kernel}

All feature constructions considered in this work are evaluated within the same
kernel-based learning framework.
Given a dataset $\{(x_i, y_i)\}_{i=1}^N$, classical preprocessing is first applied
to obtain a transformed representation for each input, which is subsequently
mapped to a quantum state using the chosen feature construction.
Kernel values are then obtained by estimating pairwise state overlaps.

For a given training set, we construct a kernel matrix
$K \in \mathbb{R}^{N \times N}$, where each entry $K_{ij}$ corresponds to the
estimated overlap between the quantum states associated with $x_i$ and $x_j$.
To mitigate trivial correlations and improve numerical stability, kernel matrices
are centered prior to downstream evaluation.
Centered kernels are used consistently for all diagnostic metrics and learning
tasks.

Learning performance is assessed using a support vector machine (SVM) with a
precomputed kernel.
Train--test splits are generated using stratified sampling to preserve class
balance, and all reported metrics are averaged over multiple random seeds.
This protocol ensures that observed performance differences reflect systematic
effects of the feature construction rather than artifacts of a particular data
split.

In addition to classification accuracy and macro-averaged F1 score, we report
kernel-level diagnostics that provide insight into representation quality.
Specifically, we compute kernel alignment with the ideal label kernel, as well as
the difference between within-class and between-class similarities.
These quantities characterize how well the induced kernel geometry reflects class
structure independently of the classifier.

To analyze robustness under noise, kernel construction and evaluation are repeated
across a range of noise levels.
Noise is injected at the data level prior to preprocessing, ensuring that all
methods are exposed to identical perturbations.
This design isolates the effect of the feature construction on robustness and
allows degradation trends to be compared in a controlled and reproducible manner.

Unless stated otherwise, the same evaluation protocol is used for all classical
front-ends and quantum feature maps, enabling a fair and systematic comparison.
\subsection{Experimental setup}
\label{subsec:exp_setup}

Our empirical evaluation spans 20 widely used image-like classification benchmarks covering heterogeneous visual regimes (e.g., digits, natural images, faces, and medical imagery). These datasets are commonly employed in both classical and quantum machine learning studies, enabling reproducible and externally comparable robustness assessments \ref{supp:github}. To make results comparable across datasets with different native resolutions and formats, all inputs are converted to grayscale when applicable, rescaled to a common $32\times 32$ resolution, and normalized to $[0,1]$ prior to feature construction. For each noise level we subsample $N=150$ examples using class-balanced sampling to prevent spurious shifts in effective class priors. Dataset metadata (classes and characteristics) are reported in Appendix~\ref{app:experimental_details} (Table~\ref{tab:datasets_overview}).

Robustness is assessed under controlled additive Gaussian corruption applied to
normalized inputs,
\begin{equation}
x \mapsto \mathrm{clip}_{[0,1]}(x + \eta), \quad \eta \sim \mathcal{N}(0,\sigma^2),
\label{eq:noise_model}
\end{equation}
with $\sigma \in \{0.00, 0.025, 0.05, 0.075, 0.10, 0.125, 0.15, 0.20\}$. The upper bound $\sigma=0.20$ corresponds to perturbations with variance $0.04$ in the $[0,1]$ domain, representing a high-noise regime where structure is significantly degraded but not fully destroyed, thereby enabling slope-based robustness comparisons before universal chance-level collapse.

To isolate the role of the feature construction, we evaluate a family of quantum-kernel and classical-kernel pipelines that share the same downstream classifier (SVM with a precomputed kernel) and differ only in the classical front-end that generates the $m$-dimensional representation feeding the quantum embedding. We consider three standard and widely adopted front-end constructions in machine learning: a DFT-based spectral selection, principal component analysis (PCA), and random projections (RP). These represent, respectively, frequency-domain feature extraction, variance- maximizing linear compression, and structure-agnostic dimensionality reduction. To ensure a controlled comparison, we tie the retained feature dimension to the quantum state dimension, $m = 2^n$, so that all variants operate at identical effective dimensionality for fixed circuit size. DFT features are selected using a low-frequency radial criterion in 2D Fourier space, while PCA and RP are constructed to match the same $m$.

Kernel entries are estimated via quantum overlap estimation using a fixed number of circuit repetitions (shots), i.e., repeated projective measurements used to estimate expectation values under finite sampling. This fixed shot budget ensures comparable statistical precision across all variants, producing $K_{\mathrm{tr}}$ and $K_{\mathrm{te,tr}}$; an SVM is trained using the precomputed kernel. All normalization statistics and any diagnostic kernel quantities are computed strictly on training data to prevent leakage \cite{cristianini_kernel-target_2001}. We use stratified train--test splits with a $30\%$ test fraction, repeated over five random seeds; performance metrics are averaged across seeds.

A central difficulty in cross-method robustness comparisons is that each dataset admits multiple valid configurations (e.g., different $n$ and associated $m$, and possibly different budgets). To provide a fair comparison while allowing each method family to operate in a realistic tuned regime, we adopt a clean-data hyperparameter selection protocol. Specifically, for each dataset $d$ and each method group $g$, we select the configuration that maximizes mean accuracy at a reference noise level $\sigma_0$ (here $\sigma_0=0$), and then freeze that choice for all $\sigma\in\mathcal{S}$. This mirrors standard practice in which hyperparameters are tuned on clean (or lightly corrupted) data and robustness is assessed out-of-distribution, while explicitly avoiding noise-dependent retuning.

In particular, the group denoted QK\_DFT corresponds to the SPE construction introduced in Sec.~\ref{subsec:methods_spe}. Under this protocol we compare five groups within a single unified statistical framework: QK\_DFT (i.e., SPE), QK\_PCA, and QK\_RP—three quantum kernels sharing the same diagonal embedding but differing in their classical front-end—and two classical baselines, SVM\_Linear and SVM\_RBF, evaluated under the same preprocessing choice used in the classical branch. Robustness is quantified using a dataset fixed-effects regression model with group-specific degradation slopes. Taking QK\_DFT as the reference group, we fit
\begin{equation}
    \mathrm{Acc}_{i} = \alpha_{d(i)} + \beta_{\text{QK-DFT}} \sigma_i + \sum_{g \neq \text{QK-DFT}} \delta_g \, \sigma_i \mathbf{1}_{g(i)} + \varepsilon_i,
\end{equation}
so that each group has slope $\beta_g=\beta_{\text{QK-DFT}}+\delta_g$. Inference is performed with a wild cluster bootstrap using Rademacher weights at the dataset level ($B=4000$ replications), which preserves within-dataset dependence and yields robust confidence intervals for slope differences. For visualization, we report accuracy--vs--noise curves after sigma$_0$ selection, together with bootstrap median curves and $95\%$ bands to reflect between-dataset variability. 

Finally, to validate practical executability, we conduct a hardware micro-benchmark on IBM Quantum backends using a SWAP-test overlap estimation task (Sec.~\ref{subsec:results_hardware}). The objective is not end-to-end classification on hardware, but to assess overlap-estimation stability under realistic compilation and gate noise.

\section{Results}
\label{sec:results}
The primary objective of this study is to characterize how different feature constructions degrade as input noise increases, thereby providing a robustness-centered evaluation rather than a peak-accuracy comparison. In realistic machine learning pipelines—particularly in hybrid quantum--classical settings—data corruption is unavoidable, and performance stability under perturbations is often more relevant than peak clean accuracy. Consequently, evaluating methods at a single noise level can be misleading: a model that performs strongly at $\sigma=0$ but deteriorates rapidly may be less desirable than a model with slightly lower clean accuracy but substantially greater robustness.

In this context, comparing methods solely at a fixed noise level can be misleading. A method that achieves high accuracy at low noise but deteriorates rapidly may be less desirable than a method that exhibits slightly lower peak performance but degrades more gracefully. Accordingly, we adopt a noise-aware evaluation strategy in which the noise level $\sigma$ is treated as a controlled experimental variable, and performance is analyzed as a function of increasing noise. Note that the underlying benchmark datasets already contain inherent acquisition and preprocessing noise; the added Gaussian perturbation therefore represents controlled incremental corruption rather than a transition from perfectly clean data. This evaluation strategy is consistent with established corruption-robustness benchmarks in computer vision. As surveyed in ~\cite{wang2024surveyrobustnesscomputervision}, corruption benchmarks such as CIFAR-C and ImageNet-C, originally introduced by Hendrycks and Dietterich~\cite{hendrycks2019benchmarkingneuralnetworkrobustness}, systematically apply controlled levels of common perturbations—including Gaussian noise, blur, and digital distortions—to clean datasets in order to quantify performance degradation under distribution shift. 

All results in this section are constructed under a unified experimental protocol described in Sec.~\ref{subsec:exp_setup}. In particular, hyperparameters are selected at a clean reference noise level $\sigma_0=0$ separately for each dataset and method group, and then frozen for all higher noise levels. This prevents noise-dependent retuning and mirrors standard practice in which models are tuned on clean data and evaluated under distribution shift. Robustness is then quantified through a dataset fixed-effects regression model with group-specific degradation slopes, using wild cluster bootstrap inference at the dataset level.

We jointly compare five method groups within a single regression framework: QK\_DFT, QK\_PCA, QK\_RP, SVM\_Linear, and SVM\_RBF. QK\_DFT (spectral preprocessing followed by diagonal quantum encoding) is treated as the reference group. This unified modeling approach ensures that all slope estimates and confidence intervals arise from the same statistical experiment, avoiding inconsistencies that can arise from method-specific aggregation rules.

Throughout this section, we present complementary views of robustness. First, we report global degradation slopes obtained from the fixed-effects regression, which quantify accuracy loss per unit increase in $\sigma$. These slopes provide a compact and statistically grounded summary of robustness differences across methods. Second, we present accuracy-versus-noise curves with bootstrap confidence bands, offering a more granular visualization of degradation trajectories. Together, these analyses provide both quantitative and descriptive evidence regarding relative robustness.

Importantly, our analysis focuses on regimes in which performance degradation remains informative, rather than on asymptotic collapse scenarios. The emphasis is therefore on comparative stability across increasing corruption levels, rather than on characterizing trivial high-noise behavior. In this way, the results directly address the central question of this work: whether spectral preprocessing combined with diagonal quantum encoding yields a more favorable degradation profile than alternative classical or quantum feature constructions under controlled noise.

\subsection{Performance Evolution under Noise}
\label{subsec:typical_vs_upper}

\begin{figure}[t]
\centering
\includegraphics[width=0.48\linewidth]{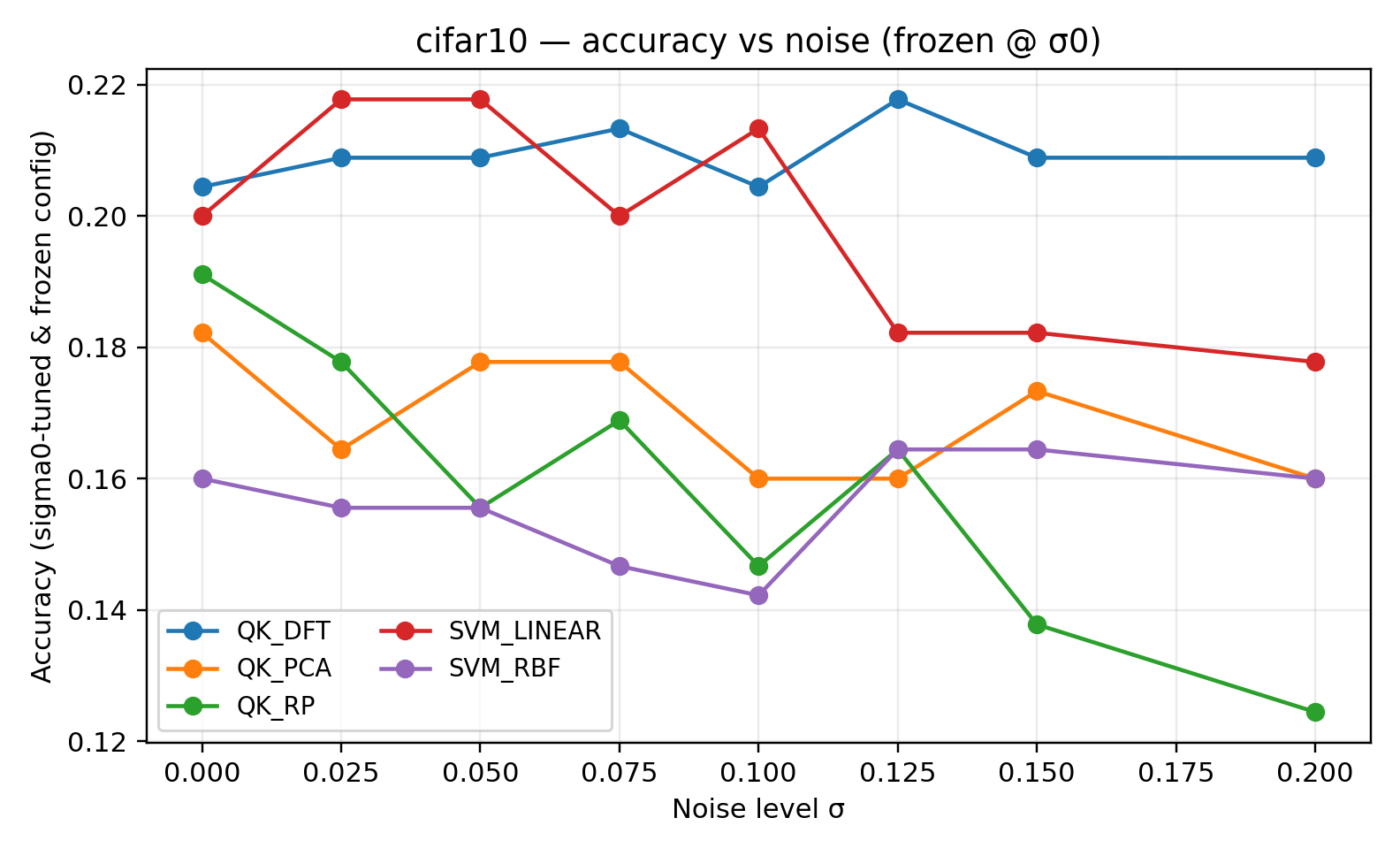}
\includegraphics[width=0.48\linewidth]{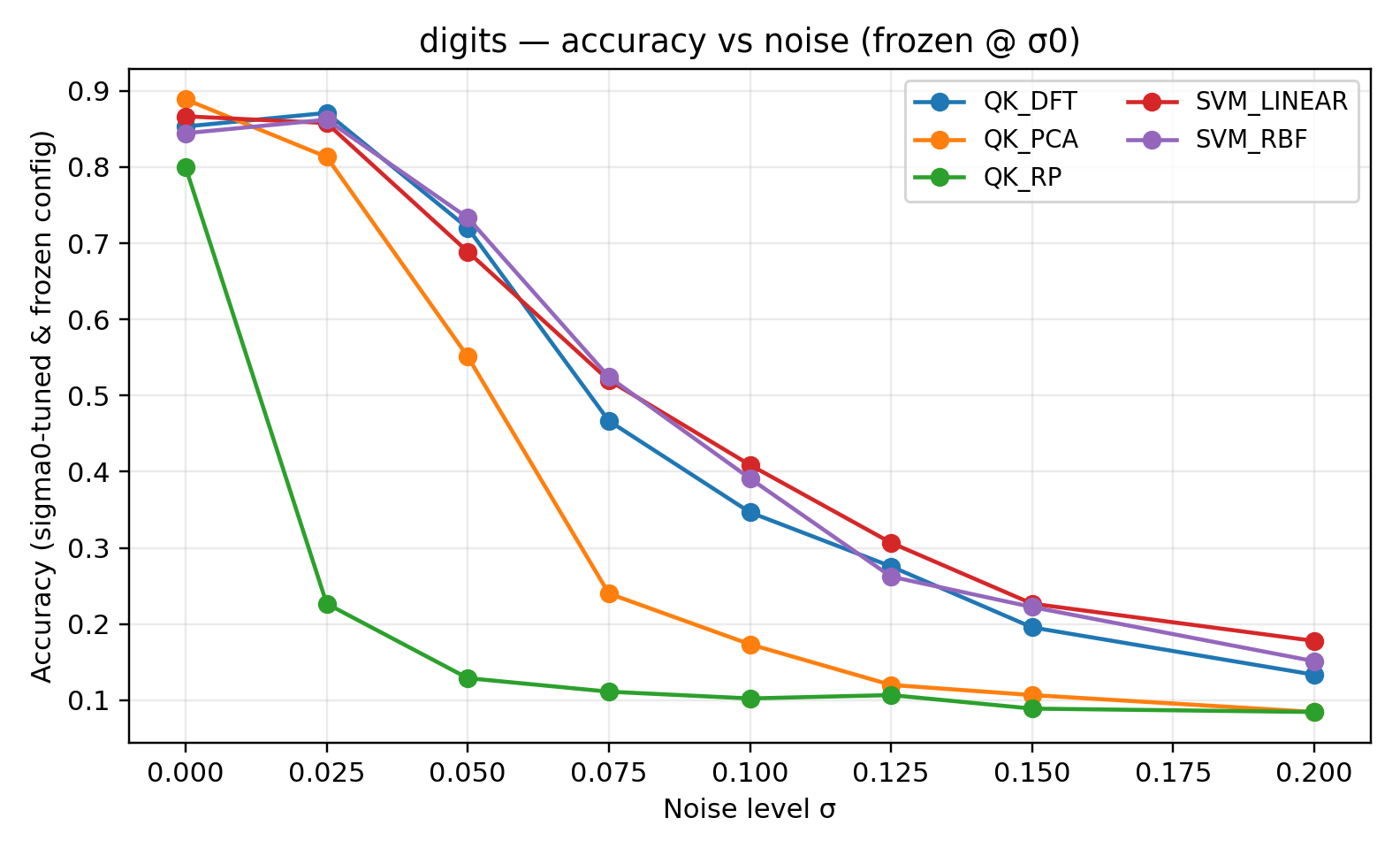}

\includegraphics[width=0.48\linewidth]{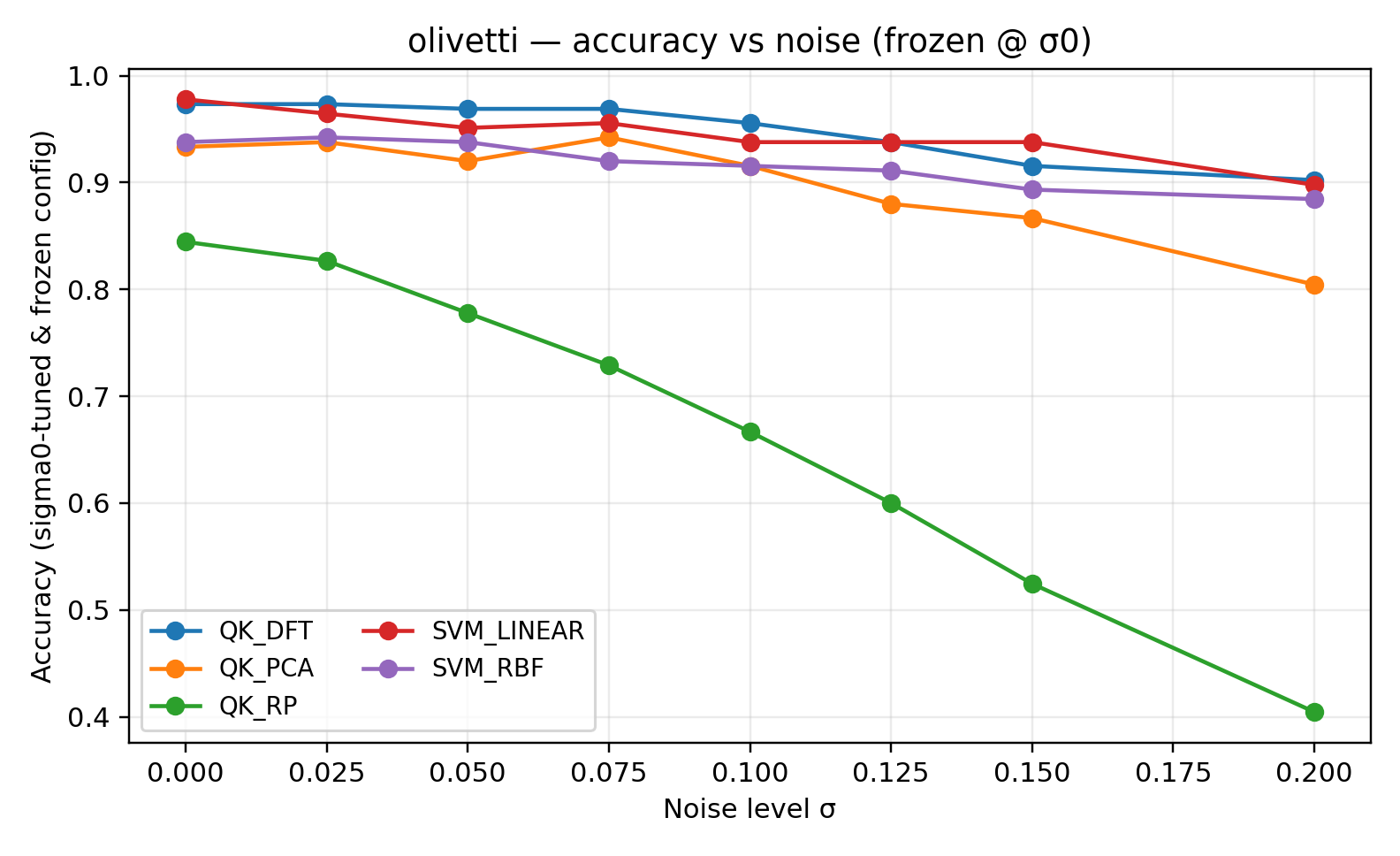}
\includegraphics[width=0.48\linewidth]{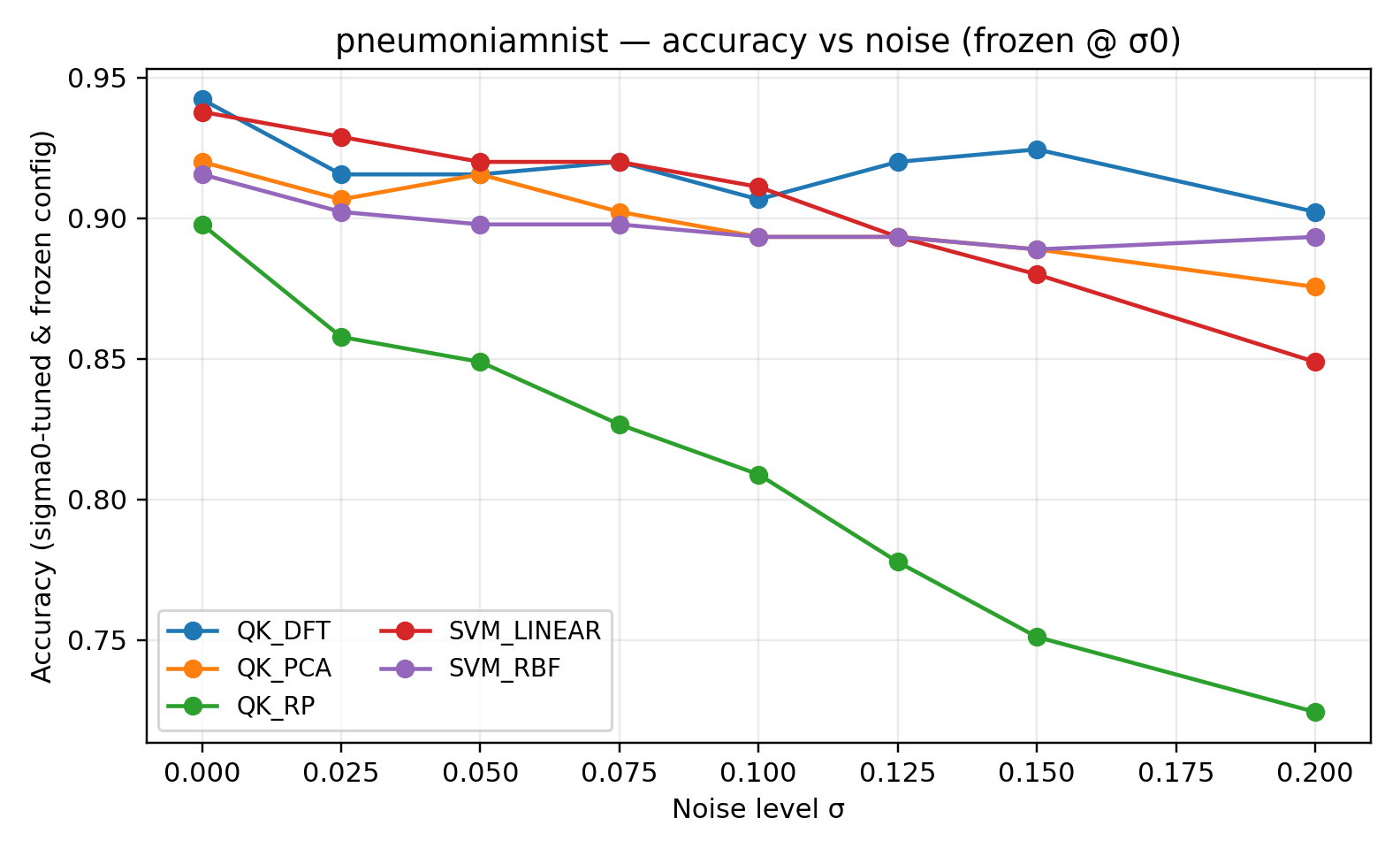}
\caption{Accuracy vs noise level $\sigma$ for four representative datasets under sigma$_0$-frozen configuration selection.}
\label{fig:curves_representative}
\end{figure}

We first analyze how classification accuracy evolves as a function of increasing additive Gaussian noise under the clean-data hyperparameter selection protocol described in Sec.~\ref{subsec:exp_setup}. The visual evolution of the corrupted inputs corresponding to increasing $\sigma$ is illustrated in Appendix~\ref{app:samples_corruption} (Fig.~\ref{fig:noise_examples}). As the noise standard deviation $\sigma$ increases, spatial structure progressively degrades, eventually approaching visually uninformative patterns. The quantitative impact of this degradation on classification performance is shown in Fig.~\ref{fig:curves_representative}.

Figure~\ref{fig:curves_representative} reports accuracy--vs--noise curves for four representative datasets spanning natural images, handwritten digits, face recognition, and medical imaging. 
For each dataset and each method group, the configuration achieving the highest mean accuracy at $\sigma_0=0$ is selected and then \emph{frozen} for all subsequent noise levels. 
This mirrors standard practice in which hyperparameters are tuned on clean data and robustness is evaluated out-of-distribution without noise-dependent retuning.

Across datasets, a consistent qualitative pattern emerges. 
Among the quantum variants, QK\_DFT exhibits a systematically milder degradation compared to QK\_PCA and especially QK\_RP, whose slopes are visibly steeper as $\sigma$ increases. 
The classical baselines (SVM\_Linear and SVM\_RBF) display intermediate behavior, with degradation patterns that vary across datasets but do not systematically outperform QK\_DFT in the moderate-noise regime.

\begin{figure}[t]
    \centering
    \includegraphics[width=0.65\linewidth]{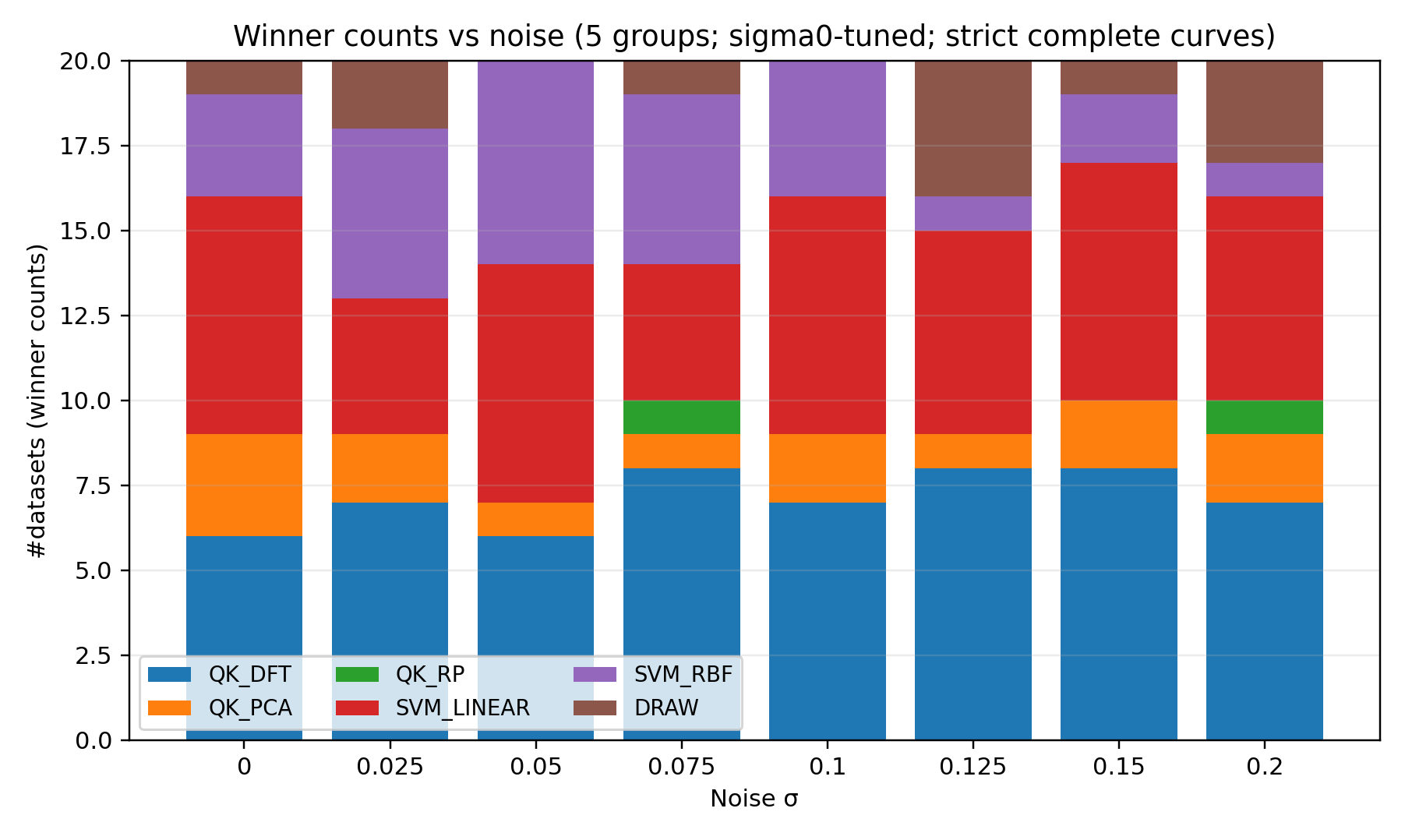}
    \caption{Number of datasets (out of 20) for which each method achieves the highest accuracy at each noise level $\sigma$, after clean-data configuration selection and freezing.}
    \label{fig:winners_vs_sigma}
\end{figure}

To assess this behavior at scale, we aggregate winner counts across all 20 datasets at each noise level using the frozen-configuration protocol. 
Figure~\ref{fig:winners_vs_sigma} reports the number of datasets for which each method achieves the highest accuracy at each $\sigma$.

QK\_DFT achieves the largest number of wins across most noise levels, particularly in the intermediate corruption regime ($0.05 \le \sigma \le 0.15$), where structural information remains partially preserved but fine details are increasingly distorted. In this range, QK\_DFT attains between 7 and 8 wins out of 20 datasets, consistently matching or exceeding the performance of the classical baselines. In contrast, QK\_PCA and especially QK\_RP exhibit markedly fewer wins across the spectrum, reflecting their steeper degradation patterns observed in the per-dataset curves. While SVM\_Linear and SVM\_RBF occasionally achieve competitive or even leading performance at specific noise levels (notably at $\sigma=0$ and $\sigma=0.05$), no consistent dominance over QK\_DFT emerges as noise increases.

Taken together, these aggregated winner counts indicate that the Fourier-based quantum construction (QK\_DFT) maintains a comparatively stable competitive position under corruption. Rather than relying on isolated peak accuracies, the advantage manifests as sustained top performance across a broad noise range, suggesting that the combination of spectral preprocessing and diagonal phase embedding yields a robustness profile that is not systematically surpassed by alternative quantum front-ends or classical kernel baselines under the same clean-data tuning protocol.

\subsection{Noise robustness analysis}
\label{subsec:noise_robustness_all}

We now quantify robustness using the regression framework introduced in Sec.~\ref{subsec:exp_setup}, applied to the sigma$_0$-tuned configurations of all five method groups simultaneously. Robustness is measured by the degradation slope $\beta_g$, defined as the change in accuracy per unit increase of noise level $\sigma$.

\begin{figure}[t]
    \centering
    \begin{subfigure}[t]{0.48\linewidth}
        \centering
        \includegraphics[width=\linewidth]{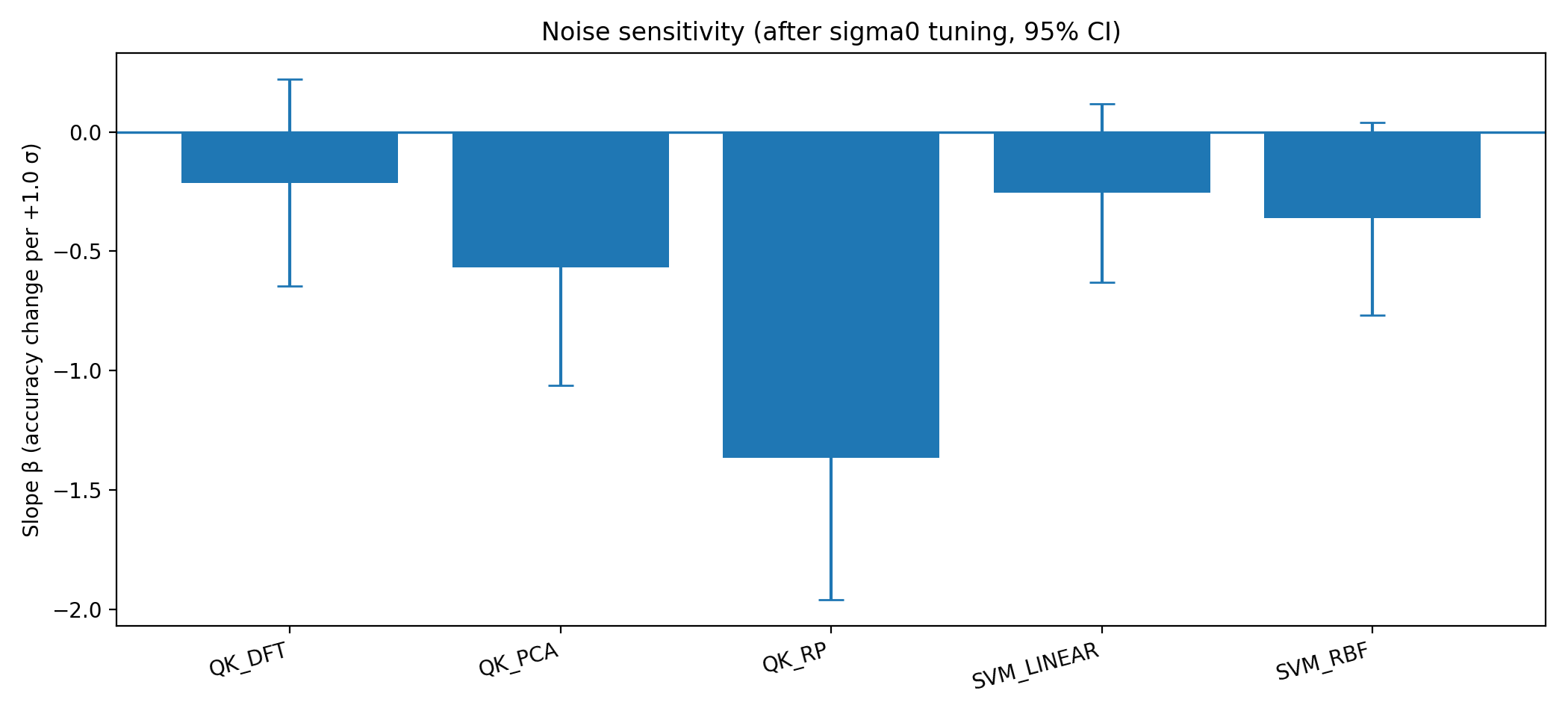}
        \caption{Noise sensitivity by method. Bars show the estimated degradation slopes $\beta$ (accuracy change per unit increase in noise $\sigma$), with 95\% wild cluster bootstrap confidence intervals (clustered by dataset, $B=4000$).}
        \label{fig:noise_betas_all}
    \end{subfigure}
    \hfill
    \begin{subfigure}[t]{0.48\linewidth}
        \centering
        \includegraphics[width=\linewidth]{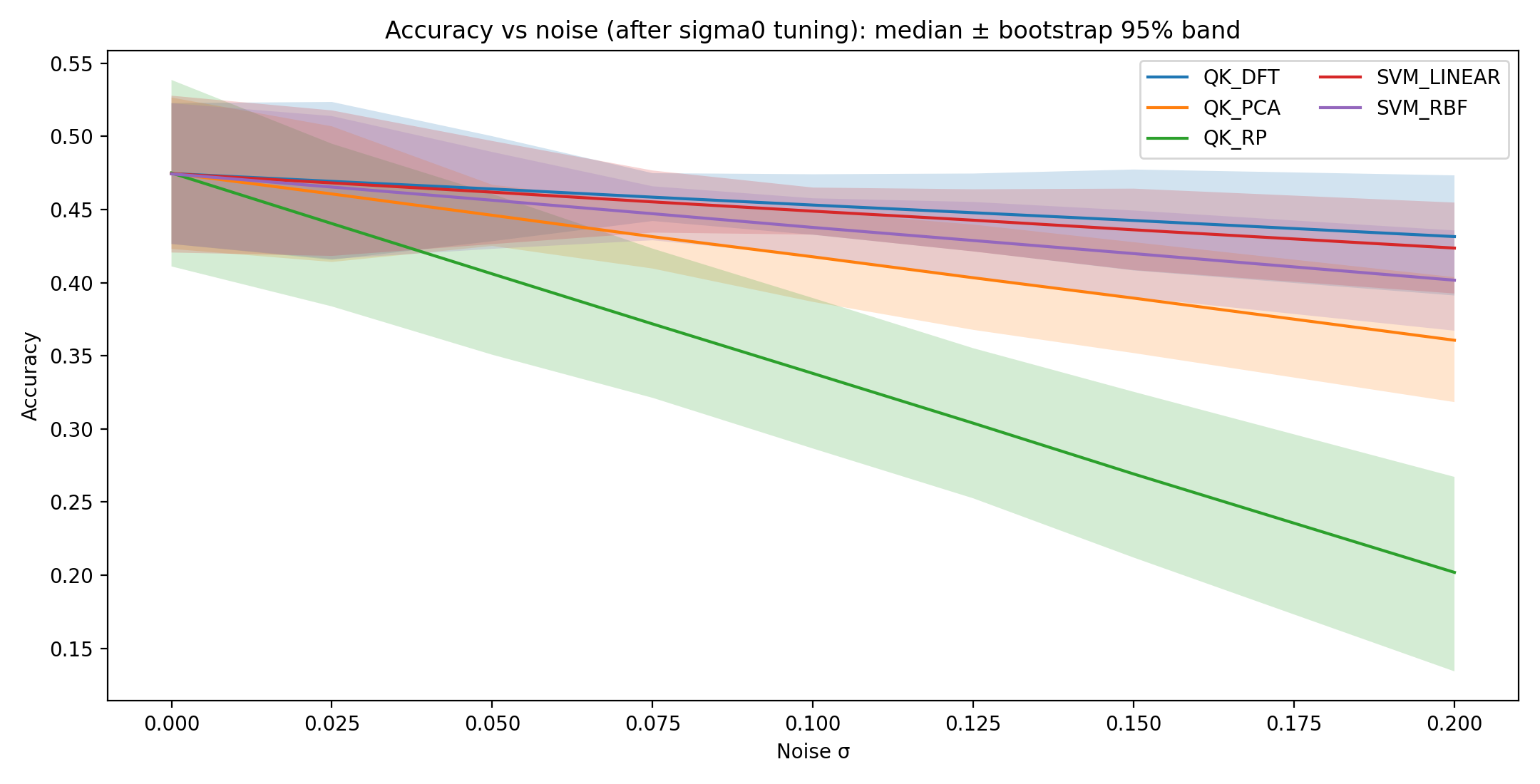}
        \caption{Global accuracy vs noise level $\sigma$ aggregated across complete dataset--configuration curves. Lines show the bootstrap median and shaded regions indicate the 95\% wild cluster bootstrap bands (clustered by dataset, $B=4000$).}
        \label{fig:noise_curves_all}
    \end{subfigure}
    \caption{Global robustness analysis. (a) Degradation slopes with bootstrap confidence intervals. (b) Aggregated accuracy curves with 95\% bootstrap bands.}
    \label{fig:bootstrap_combined}
\end{figure}

Figure~\ref{fig:bootstrap_combined} summarizes the results. Panel (a) reports the estimated slopes with 95\% wild cluster bootstrap confidence intervals (clustered at the dataset level, $B=4000$), and Panel (b) shows the corresponding aggregated accuracy curves with bootstrap bands. Numerical estimates are provided in Table~\ref{tab:noise_slopes_all}.

\begin{table*}[t]
\centering
\caption{Noise robustness slopes estimated via dataset fixed-effects regression with wild cluster bootstrap. Slopes correspond to accuracy change per unit increase in noise $\sigma$. QK-DFT is the reference category.}
\label{tab:noise_slopes_all}
\begin{tabular}{lcccc}
\toprule
Parameter & Estimate & CI$_{95\%}$ Low & CI$_{95\%}$ High & $\Pr(\beta<0)$ \\
\midrule
$\beta_{\text{QK-DFT}}$      & -0.214 & -0.645 &  0.222 & 0.801 \\
$\beta_{\text{QK-PCA}}$      & -0.568 & -1.063 & -0.075 & 0.995 \\
$\Delta(\text{QK-PCA}-\text{QK-DFT})$ & -0.354 & -0.524 & -0.188 & 1.000 \\
$\beta_{\text{QK-RP}}$       & -1.367 & -1.962 & -0.760 & 1.000 \\
$\Delta(\text{QK-RP}-\text{QK-DFT})$  & -1.153 & -1.592 & -0.707 & 1.000 \\
$\beta_{\text{SVM-Linear}}$  & -0.253 & -0.630 &  0.118 & 0.867 \\
$\Delta(\text{SVM-Linear}-\text{QK-DFT})$ & -0.040 & -0.179 &  0.093 & 0.714 \\
$\beta_{\text{SVM-RBF}}$     & -0.362 & -0.769 &  0.041 & 0.948 \\
$\Delta(\text{SVM-RBF}-\text{QK-DFT})$ & -0.149 & -0.289 & -0.011 & 0.986 \\
\bottomrule
\end{tabular}
\end{table*}

\begin{align*}
\beta_{\text{QK-DFT}} &= -0.214, \\
\beta_{\text{QK-PCA}} &= -0.568, \\
\beta_{\text{QK-RP}}  &= -1.367, \\
\beta_{\text{SVM-Linear}} &= -0.253, \\
\beta_{\text{SVM-RBF}} &= -0.362.
\end{align*}

Within the quantum family, DFT-based encoding exhibits the smallest degradation rate, while PCA and especially random projections deteriorate substantially faster. The differences relative to QK-DFT are:

\begin{align*}
\Delta(\text{QK-PCA} - \text{QK-DFT}) &= -0.354, \\
\Delta(\text{QK-RP} - \text{QK-DFT})  &= -1.153,
\end{align*}

with bootstrap confidence intervals strictly below zero, indicating a statistically robust advantage of spectral preprocessing within quantum models.

When compared to classical baselines, QK-DFT displays a degradation rate comparable to linear SVM and more favorable than RBF SVM. In particular:

\begin{align*}
\Delta(\text{SVM-Linear} - \text{QK-DFT}) &= -0.040, \\
\Delta(\text{SVM-RBF} - \text{QK-DFT})    &= -0.149.
\end{align*}

The linear SVM difference is not statistically significant, whereas the RBF SVM exhibits significantly stronger degradation under noise. 

Taken together, these results show that robustness is strongly governed by preprocessing alignment: spectral (DFT) representations yield consistently smaller noise sensitivity both in quantum and classical settings. However, the diagonal quantum feature map preserves this stability and avoids the stronger degradation observed in nonlinear classical kernels under the same spectral features.

\subsection{Hardware validation: overlap estimation under realistic noise}
\label{subsec:results_hardware}

To complement the simulation-based robustness analysis, we assess the practical
executability of SPE—denoted QK-DFT in the unified regression framework—on current superconducting quantum hardware.
Rather than performing end-to-end classification on hardware, which would confound
kernel behavior with training variance and limited shot budgets, we isolate the
fundamental primitive underlying the quantum kernel: state overlap estimation.
Given two real-valued inputs $(x,y)$ with prescribed cosine similarity $\rho$, we
prepare the corresponding quantum states $\ket{\psi(x)}$ and $\ket{\psi(y)}$
using the SPE construction (QK-DFT) or a low-depth angle-encoding baseline.
The overlap is estimated via a SWAP test, producing a hardware probability
$p_0^{\mathrm{hw}}$, which we compare against its expected reference value
$p_0^{\mathrm{exp}}$ obtained from analytic calculation (SPE) or noiseless
statevector simulation (angle baseline). Hardware fidelity is summarized through the absolute deviation
\[ \lvert \Delta p_0 \rvert = \lvert p_0^{\mathrm{hw}} - p_0^{\mathrm{exp}} \rvert,\]
which directly measures how accurately the device reproduces the theoretical kernel
overlap.

Experiments were executed on two distinct IBM Quantum backends,
\texttt{ibm\_fez} and \texttt{ibm\_marrakesh}, using identical circuit constructions,
transpilation strategies, and shot budgets. Results are averaged across backends,
multiple similarity settings, and repeated runs in order to mitigate backend-specific
calibration fluctuations and transient noise effects.

\begin{figure}[t]
    \centering
    \includegraphics[width=0.75\linewidth]{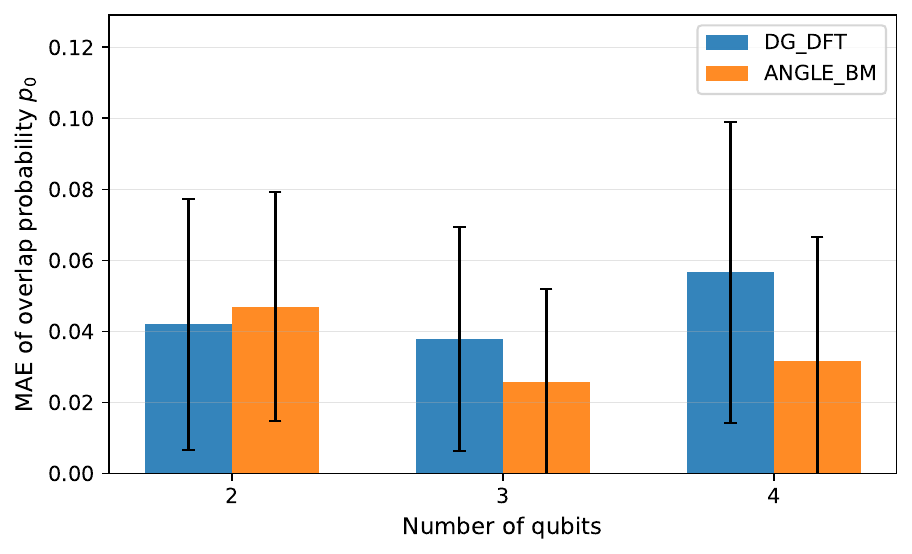}
    \caption{
    Hardware overlap-estimation error across qubit sizes.
    Mean absolute error (MAE) of
    $\lvert \Delta p_0 \rvert = \lvert p_0^{\mathrm{hw}} - p_0^{\mathrm{exp}} \rvert$
    for SPE (DFT + DiagonalGate) and a low-depth angle-encoding baseline, averaged over
    two IBM Quantum backends (\texttt{ibm\_fez}, \texttt{ibm\_marrakesh}),
    similarity values, and repeated runs.
    Error bars indicate variability across measurement instances.
    }
    \label{fig:hw_mae}
\end{figure}

Figure~\ref{fig:hw_mae} reports the mean absolute error as a function of qubit count.
Across all tested regimes ($n=2,3,4$), deviations remain moderate and do not exhibit
catastrophic growth with increasing system size. Although the relative ordering between
SPE and the angle baseline varies with qubit count, both constructions produce stable
and non-degenerate overlap estimates. Importantly, the diagonal phase embedding does
not display instability or collapse as circuit depth increases, despite relying on
structured phase accumulation across computational basis states.

These results provide empirical evidence that the structural properties exploited by
SPE—namely diagonal phase encoding and low entangling depth—translate into hardware
executability on present-day superconducting devices. Combined with the large-scale
simulation analysis, this hardware validation supports the practical viability of
SPE as a kernel-construction mechanism compatible with current NISQ architectures.

\section{Discussion}
\label{sec:discussion}

This work reframes the evaluation of quantum kernel constructions from peak clean-data performance toward controlled degradation under distributional corruption. 
Rather than asking whether a given encoding achieves the highest accuracy at $\sigma=0$, we quantify how performance decays as additive noise increases, and compare degradation rates across quantum and classical pipelines under matched tuning conditions. 
This perspective is consistent with recent arguments that quantum advantage should not be framed as a binary outcome, but assessed through stability, structural alignment, and feasibility under realistic NISQ constraints \cite{schuld_is_2022, jerbi_quantum_2023}.

Methodologically, the study integrates three elements into a single robustness framework: 
(i) clean-data hyperparameter selection with frozen configurations, 
(ii) dataset fixed-effects regression with group-specific noise slopes, and 
(iii) wild cluster bootstrap inference at the dataset level. 
This design prevents optimistic noise-dependent retuning and avoids overconfident inference due to repeated measurements within datasets. 
In contrast to descriptive degradation plots alone, the regression model yields interpretable slope coefficients $\beta_g$ representing accuracy loss per unit noise increase, enabling direct cross-family comparison.

Across the five model families considered (QK-DFT, QK-PCA, QK-RP, SVM-Linear, SVM-RBF), a consistent structure emerges. 
Within the quantum family, QK-DFT exhibits the smallest degradation magnitude, followed by QK-PCA and then QK-RP. 
The slope differences between QK-PCA and QK-DFT, and especially between QK-RP and QK-DFT, are supported by bootstrap intervals excluding zero, indicating a statistically robust advantage of Fourier-based preprocessing within the diagonal quantum embedding. 
By contrast, QK-PCA and QK-RP degrade substantially faster as corruption increases.

The particularly steep degradation observed for random projections is consistent with classical dimensionality-reduction theory. 
While random projections preserve pairwise distances in expectation under Johnson- Lindenstrauss guarantees \cite{achlioptas_database-friendly_2003, li_very_2006}, they do not exploit data-dependent or task-aligned structure. 
Under additive noise, this structure-agnostic property provides no preferential preservation of discriminative directions, leading to accelerated collapse relative to variance-based (PCA) or frequency-aware (DFT) representations. 
PCA, by construction, concentrates maximal variance directions \cite{shlens_tutorial_2014, jolliffe_principal_2016}, which partially explains its intermediate robustness profile.

The favorable behavior of QK-DFT can be interpreted through the interaction between spectral preprocessing and diagonal unitary feature maps. 
Fourier-domain representations have repeatedly been shown to act as an inductive bias in classical learning, improving robustness and expressivity in noisy image and signal settings \cite{yin_fourier_2019, xu_learning_2020, stuchi_frequency_2020, tancik_fourier_2020}. 
Injecting this spectral structure into a phase-only diagonal quantum embedding preserves interference-based relational information without increasing circuit depth. 
Diagonal encodings are known to be hardware-compatible and shallow relative to more entanglement-heavy constructions \cite{PhysRevA.103.032430, havlicek_supervised_2019}, and recent analyses emphasize the importance of structure-aware and symmetry-respecting feature maps for avoiding kernel concentration and expressivity pathologies \cite{glick_covariant_2024, kuebler_inductive_bias_quantum_kernels_2021}. 
Our results suggest that the combination of Fourier structure and diagonal phase encoding provides a favorable compromise between expressivity and stability under corruption.

Importantly, the comparison with classical SVM baselines clarifies that robustness gains are not purely a consequence of preprocessing. 
When evaluated under identical clean-data tuning, QK-DFT exhibits a degradation rate comparable to linear SVM and more stable than RBF SVM under the same spectral preprocessing. 
This indicates that robustness is jointly governed by representation alignment and the geometry induced by the kernel map, rather than by the classical front-end alone.

The hardware validation further supports the practical viability of this construction. 
While QK-DFT does not uniformly outperform a low-depth angle-encoding baseline in absolute overlap-estimation error, it remains numerically stable across qubit counts and backends. 
Given the well-known sensitivity of hardware experiments to calibration, transpilation, and backend-specific noise profiles \cite{hubregtsen_training_2022, wu_application_2021}, the key observation is that the diagonal phase construction does not exhibit instability or collapse under realistic device noise. 
This operational stability is consistent with its shallow, diagonal structure and supports its suitability for near-term quantum kernel pipelines.

Overall, the findings position Spectral Phase Encoding as a structurally aligned quantum feature construction whose advantage lies not in universal peak accuracy, but in slower degradation under corruption and in statistically supported slope improvements within the quantum family. 
The results also highlight that robustness is fundamentally domain-conditional: spectral alignment benefits datasets exhibiting exploitable frequency structure, while offering limited gains when such structure is weak. 
This nuanced picture aligns with the broader shift in quantum machine learning toward domain-aware, noise-conscious algorithm design, where progress is likely to depend less on universal advantage claims and more on identifying structurally compatible problem classes.

\section{Conclusion}
\label{sec:conclusion}

We introduced Spectral Phase Encoding (SPE), a hybrid quantum feature construction that combines a discrete Fourier transform (DFT) front-end with a phase-only diagonal quantum embedding. Within the unified experimental framework, this construction is instantiated as QK-DFT and evaluated alongside alternative quantum front-ends (QK-PCA, QK-RP) and classical SVM baselines under a common clean-data tuning protocol.

Rather than focusing on peak accuracy at zero noise, the central objective of this study was to characterize controlled degradation under additive corruption. By combining clean-data hyperparameter selection, dataset fixed-effects regression, and wild cluster bootstrap inference, we quantified robustness through interpretable degradation slopes across 20 heterogeneous datasets.

The results consistently show that within the quantum family, QK-DFT exhibits the smallest magnitude degradation under increasing noise, followed by QK-PCA and QK-RP. The slope differences relative to QK-DFT are statistically supported by bootstrap intervals excluding zero, indicating a robust advantage of spectral preprocessing within a diagonal quantum embedding. Importantly, when compared to classical SVM baselines under identical tuning conditions, QK-DFT displays a degradation rate comparable to linear SVM and more stable than RBF SVM, suggesting that robustness arises from the interaction between representation alignment and kernel geometry rather than from preprocessing alone.

These findings are consistent with theoretical analyses indicating that quantum kernels can suffer from expressivity collapse and concentration phenomena under noise and measurement constraints \cite{thanasilp_exponential_2024}. By injecting structured spectral information into a shallow diagonal embedding, SPE provides a construction that balances expressivity and stability without increasing circuit depth. From a hardware perspective, diagonal phase constructions remain particularly attractive on superconducting platforms, where virtual-$Z$ implementations and shallow diagonal synthesis reduce overhead and systematic error \cite{mckay_efficient_z_2017,kjaergaard_superconducting_2020,vezvaee_virtualz_2025,welch_diagonal_unitaries_2014}.

The hardware experiments further demonstrate that SPE (QK-DFT) remains operational across multiple backends and qubit counts, producing stable overlap estimates without evidence of catastrophic degradation. While not implying hardware-level dominance, this stability reinforces the practical compatibility of diagonal phase embeddings with present-day NISQ devices.

From a computational perspective, the classical front-ends considered here differ modestly in preprocessing cost. DFT-based spectral selection can be implemented in $O(d \log d)$ time via FFT routines, while PCA requires an initial covariance estimation and eigendecomposition, and random projections incur $O(dm)$ projection cost without training overhead. In our experimental regime, these classical costs remain negligible compared to the dominant expense of quantum kernel estimation under finite measurement shots. Consequently, the observed robustness differences cannot be attributed to disproportionate classical preprocessing complexity \cite{cooley_tukey_fft_1965, bingham_random_projection_2001, yang_how_to_reduce_2020}.

Overall, the advantage of SPE does not lie in universally higher clean-data accuracy, but in statistically supported slower degradation under corruption and in the existence of robust configurations when the data exhibit exploitable spectral structure. This supports the broader view that successful quantum kernel constructions require domain-compatible inductive bias rather than universal advantage claims \cite{kuebler_inductive_bias_quantum_kernels_2021}.

\subsection*{Future Directions}
\label{sec:future_directions}

The present results suggest several concrete extensions:

\begin{enumerate}
    \item \textbf{Structure-aware and symmetry-respecting extensions.}
    Integrating SPE-style spectral priors with explicitly covariant or symmetry-aware kernel constructions may further improve stability and inductive alignment. Covariant quantum kernels provide a principled mechanism for encoding invariances and have been experimentally demonstrated in superconducting settings \cite{glick_covariant_2024}.
    
    \item \textbf{Mitigating kernel concentration in larger regimes.}
    Since kernel concentration and noise-induced collapse can limit informativeness at larger scales \cite{thanasilp_exponential_2024}, mitigation strategies tailored to fidelity-based kernels remain important. Recent proposals demonstrate practical mitigation techniques for covariant fidelity kernels at larger sizes \cite{agliardi_mitigating_covariant_kernels_2025}.
    
    \item \textbf{Learnable spectral selection and hybrid transforms.}
    SPE currently relies on fixed Fourier truncation. Extensions incorporating adaptive frequency masks, learnable phase scalings, or alternative frequency transforms (e.g., DFT/DCT/wavelets) would connect this framework to the broader frequency-based inductive bias literature in modern machine learning \cite{yi2025surveyft,zhou2022fedformer,leethorp2021fnet}.
    
    \item \textbf{Scalable kernel training and alignment.}
    Incorporating trainable phase scalings or diagonal layers, optimized through kernel alignment objectives, may enhance performance while controlling circuit depth. Subsampling-based alignment strategies offer a promising route to reduce training overhead \cite{sahin_efficient_2024}.

    \item \textbf{Hardware-aware diagonal synthesis.}
    As system size grows, compilation and diagonal synthesis will increasingly determine realized depth and noise sensitivity. Leveraging optimized diagonal decompositions and virtual-$Z$-centric compilation strategies can further reduce systematic error \cite{mckay_efficient_z_2017,vezvaee_virtualz_2025,welch_diagonal_unitaries_2014}. 
    More broadly, recent analyses of Grover-based circuit constructions highlight that qubit overhead and circuit depth remain the primary bottlenecks on current NISQ hardware, even for moderately sized problem instances \cite{10.1007/978-3-032-07992-3_20}. 
    This reinforces the importance of shallow, structure-preserving encodings when aiming for practical scalability.

    \item \textbf{Bridging to classically hard diagonal feature maps.}
    Integrating spectral priors with commuting or diagonal circuit families conjectured to be classically hard to simulate may strengthen separations in kernel evaluation. Complexity-theoretic evidence for commuting circuit hardness provides motivation for such exploration \cite{bremner_commuting_2010}.
    
    \item \textbf{Application-driven evaluation under realistic constraints.}
    While several studies report promising speedups of quantum algorithms in domains such as financial optimization, practical deployment remains limited by hardware cost, calibration instability, and circuit depth constraints \cite{10.1007/978-3-032-08381-4_14}. 
    Future work should therefore evaluate encoding strategies such as SPE within domain-specific pipelines under realistic hardware and resource assumptions.
    
    \item \textbf{Error mitigation for overlap estimation.}
    Since kernel evaluation remains measurement-limited and noise-sensitive, incorporating mitigation techniques specifically designed for near-term fidelity estimation pipelines is an important practical direction \cite{agliardi_mitigating_covariant_kernels_2025}.
\end{enumerate}

\paragraph*{Supplementary information}
\label{supp:github}
The code required to reproduce all experiments, figures, and statistical analyses is publicly available at: \url{https://github.com/cloudlab-aia/kernel-comparison}. The repository includes data preprocessing scripts, kernel construction routines, evaluation pipelines, and instructions for reproducing the main results.

\paragraph*{Acknowledgements} Grant Serverless4HPC PID2023-152804OB-I00 funded by MICIU/AEI/10.13039/501100011033 and by ERDF/EU.
\FloatBarrier

\appendix
\section{Additional experimental details}
\label{app:experimental_details}
\subsection{Class selection and dataset balancing}
\label{app:class_selection}
\begin{table*}[ht]
\centering
\caption{Overview of the 20 real-world 2D datasets used in the experiments.
All datasets are converted to grayscale (when applicable), resized to $32\times32$,
and subsampled to a balanced subset as described in Sec.~\ref{subsec:exp_setup}.}
\label{tab:datasets_overview}
\begin{tabular}{lccc}
\hline
Dataset & Classes ($C$) & Domain & Source \\
\hline
BreastMNIST      & 2  & Medical imaging & \href{https://medmnist.com/}{MedMNIST} \\
CIFAR-10         & 10 & Natural images  & \href{https://www.cs.toronto.edu/~kriz/cifar.html}{CIFAR} \\
DermaMNIST       & 7  & Dermatology     & \href{https://medmnist.com/}{MedMNIST} \\
Digits           & 10 & Handwritten     & \href{https://scikit-learn.org/}{scikit-learn} \\
DTD              & 20 & Texture         & \href{https://www.robots.ox.ac.uk/~vgg/data/dtd/}{Oxford VGG} \\
EMNIST (ByClass) & 20 & Characters      & \href{https://www.nist.gov/itl/products-and-services/emnist-dataset}{NIST} \\
Fashion-MNIST    & 10 & Clothing        & \href{https://github.com/zalandoresearch/fashion-mnist}{Zalando} \\
FER2013          & 7  & Facial expr.    & \href{https://www.kaggle.com/datasets/msambare/fer2013}{Kaggle} \\
GTSRB            & 20 & Traffic signs   & \href{http://benchmark.ini.rub.de/?section=gtsrb}{GTSRB} \\
KMNIST           & 10 & Japanese char.  & \href{http://codh.rois.ac.jp/kmnist/}{KMNIST} \\
OCTMNIST         & 4  & Retinal OCT     & \href{https://medmnist.com/}{MedMNIST} \\
PneumoniaMNIST   & 2  & Chest X-ray     & \href{https://medmnist.com/}{MedMNIST} \\
RetinaMNIST      & 5  & Retinal         & \href{https://medmnist.com/}{MedMNIST} \\
STL-10           & 10 & Natural images  & \href{https://cs.stanford.edu/~acoates/stl10/}{Stanford} \\
SVHN (cropped)   & 10 & Street digits   & \href{http://ufldl.stanford.edu/housenumbers/}{SVHN} \\
USPS             & 10 & Handwritten     & \href{https://www.csie.ntu.edu.tw/~cjlin/libsvmtools/datasets/multiclass.html}{LIBSVM} \\
\hline
\end{tabular}
\end{table*}

Real-world datasets considered in this work differ substantially in their number of available classes, class imbalance, and total sample size.
To ensure a fair and reproducible comparison across datasets, we adopt a unified and deterministic class-selection rule that balances representativeness and statistical reliability. For each dataset, we first determine the total number of available classes $C_{\mathrm{avail}}$ from the full label set.
Unless explicitly overridden, the number of classes $C$ used in the experiments is then selected according to the following rule:
\begin{equation}
C = \min\!\left(
\left\lfloor \frac{N}{n_{\mathrm{spc}}} \right\rfloor,
C_{\mathrm{cap}},
C_{\mathrm{avail}}
\right),
\end{equation}
where $N$ is the total number of samples requested for the dataset, $n_{\mathrm{spc}}$ is the target minimum number of samples per class, and $C_{\mathrm{cap}}$ is a fixed upper bound on the number of classes. In all experiments, we use $n_{\mathrm{spc}} = 10$ and $C_{\mathrm{cap}} = 20$.

This rule enforces three constraints simultaneously:
(i) each class is represented by a sufficient number of samples to avoid degenerate decision boundaries,
(ii) datasets with a very large number of classes are capped to prevent overly sparse class coverage,
and (iii) the selected number of classes never exceeds the number of classes originally present in the dataset.

Once $C$ is determined, the corresponding classes are selected deterministically using a fixed random seed.
This ensures that class identities are reproducible across runs and noise levels, while avoiding any bias induced by manual class curation. Given the selected classes, a strictly class-balanced subset of $N$ samples is then drawn using stratified sampling.
\subsection{Progressive Samples Corruption}
\label{app:samples_corruption}
\begin{figure}[ht]
    \centering
    \includegraphics[width=0.60\linewidth]{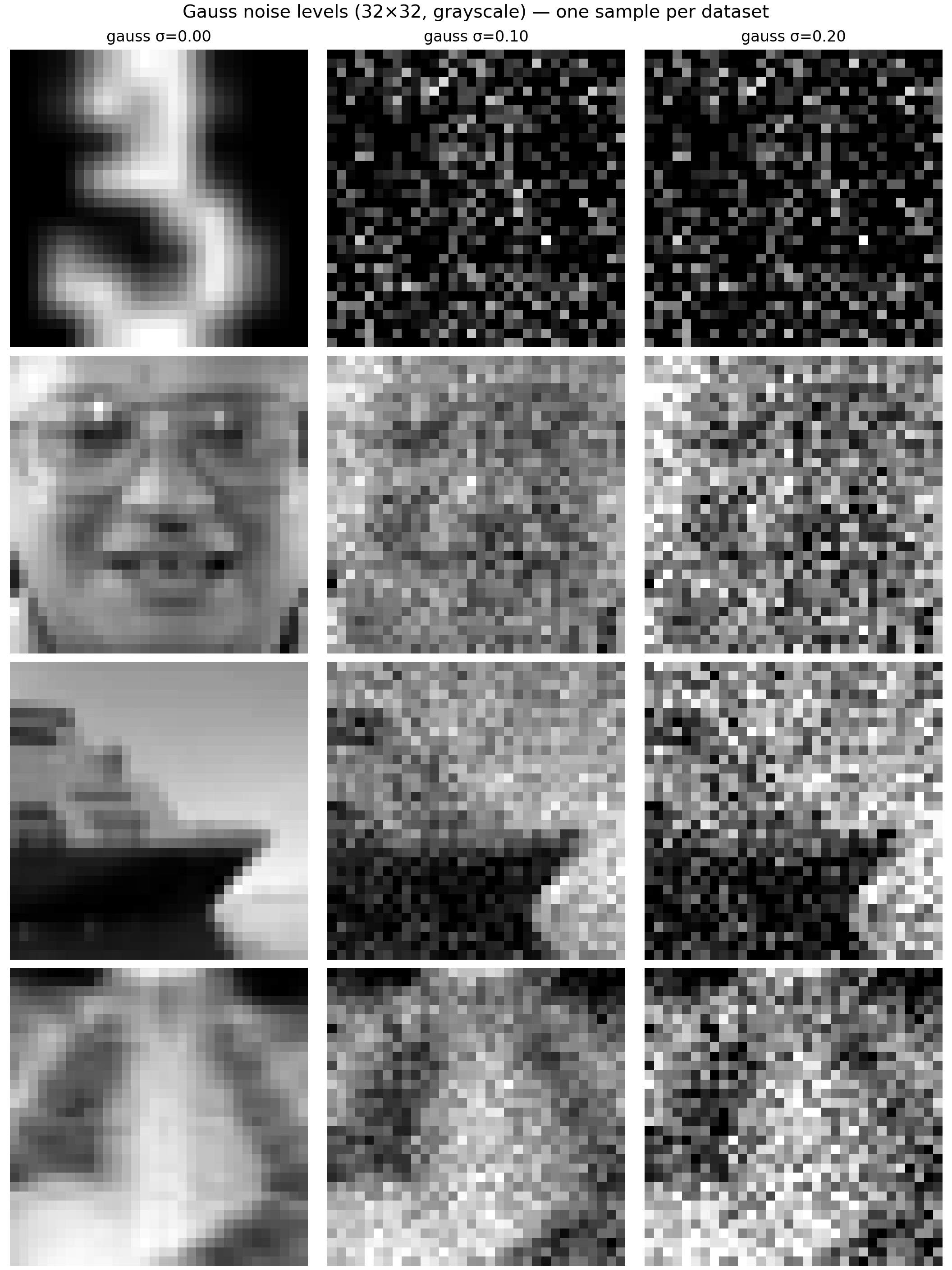}
    \caption{Representative samples under increasing Gaussian noise $\sigma$.}
    \label{fig:noise_examples}
\end{figure}


\FloatBarrier
\bibliographystyle{unsrt}
\bibliography{ref}

@misc{shlens_tutorial_2014,
	title = {A {Tutorial} on {Principal} {Component} {Analysis}},
	url = {http://arxiv.org/abs/1404.1100},
	doi = {10.48550/arXiv.1404.1100},
	publisher = {arXiv},
	author = {Shlens, Jonathon},
	month = apr,
	year = {2014},
	note = {arXiv:1404.1100 [cs]},
}

@article{jolliffe_principal_2016,
	title = {Principal component analysis: a review and recent developments},
	volume = {374},
	issn = {1364-503X},
	shorttitle = {Principal component analysis},
	url = {https://doi.org/10.1098/rsta.2015.0202},
	doi = {10.1098/rsta.2015.0202},
	number = {2065},
	urldate = {2026-01-16},
	journal = {Philosophical Transactions of the Royal Society A: Mathematical, Physical and Engineering Sciences},
	author = {Jolliffe, Ian T. and Cadima, Jorge},
	month = apr,
	year = {2016},
	pages = {20150202},
}

@misc{xu_learning_2020,
	title = {Learning in the {Frequency} {Domain}},
	url = {http://arxiv.org/abs/2002.12416},
	doi = {10.48550/arXiv.2002.12416},
	urldate = {2026-01-16},
	publisher = {arXiv},
	author = {Xu, Kai and Qin, Minghai and Sun, Fei and Wang, Yuhao and Chen, Yen-Kuang and Ren, Fengbo},
	month = mar,
	year = {2020},
	note = {arXiv:2002.12416 [cs]},
}

@inproceedings{yin_fourier_2019,
  title     = {A Fourier Perspective on Model Robustness in Computer Vision},
  booktitle = {Advances in Neural Information Processing Systems 32 (NeurIPS 2019)},
  author    = {Yin, Dong and Lopes, Raphael Gontijo and Shlens, Jonathon and Cubuk, Ekin D. and Gilmer, Justin},
  year      = {2019},
  pages     = {13276--13286},
  publisher = {Curran Associates, Inc.}
}

@misc{stuchi_frequency_2020,
	title = {Frequency learning for image classification},
	url = {http://arxiv.org/abs/2006.15476},
	doi = {10.48550/arXiv.2006.15476},
	urldate = {2026-01-16},
	publisher = {arXiv},
	author = {Stuchi, José Augusto and Boccato, Levy and Attux, Romis},
	month = jun,
	year = {2020},
	note = {arXiv:2006.15476 [cs]},
}

@inproceedings{tancik_fourier_2020,
	address = {Red Hook, NY, USA},
	series = {{NIPS} '20},
	title = {Fourier features let networks learn high frequency functions in low dimensional domains},
	isbn = {978-1-7138-2954-6},
	url = {https://dl.acm.org/doi/10.5555/3495724.3496356},
	doi = {10.5555/3495724.3496356},
	urldate = {2026-01-16},
	booktitle = {Proceedings of the 34th {International} {Conference} on {Neural} {Information} {Processing} {Systems}},
	publisher = {Curran Associates Inc.},
	author = {Tancik, Matthew and Srinivasan, Pratul P. and Mildenhall, Ben and Fridovich-Keil, Sara and Raghavan, Nithin and Singhal, Utkarsh and Ramamoorthi, Ravi and Barron, Jonathan T. and Ng, Ren},
	month = dec,
	year = {2020},
	pages = {7537--7547},
}

@misc{wang2024surveyrobustnesscomputervision,
      title={A Survey on the Robustness of Computer Vision Models against Common Corruptions}, 
      author={Shunxin Wang and Raymond Veldhuis and Christoph Brune and Nicola Strisciuglio},
      year={2024},
      eprint={2305.06024},
      archivePrefix={arXiv},
      primaryClass={cs.CV},
      url={https://arxiv.org/abs/2305.06024}, 
}

@misc{hendrycks2019benchmarkingneuralnetworkrobustness,
      title={Benchmarking Neural Network Robustness to Common Corruptions and Perturbations}, 
      author={Dan Hendrycks and Thomas Dietterich},
      year={2019},
      eprint={1903.12261},
      archivePrefix={arXiv},
      primaryClass={cs.LG},
      url={https://arxiv.org/abs/1903.12261}, 
}

@article{schuld_is_2022,
	title = {Is {Quantum} {Advantage} the {Right} {Goal} for {Quantum} {Machine} {Learning}?},
	volume = {3},
	url = {https://link.aps.org/doi/10.1103/PRXQuantum.3.030101},
	doi = {10.1103/PRXQuantum.3.030101},
	number = {3},
	urldate = {2026-01-16},
	journal = {PRX Quantum},
	publisher = {American Physical Society},
	author = {Schuld, Maria and Killoran, Nathan},
	month = jul,
	year = {2022},
	pages = {030101},
}

@article{jerbi_quantum_2023,
	title = {Quantum machine learning beyond kernel methods},
	volume = {14},
	copyright = {2023 The Author(s)},
	issn = {2041-1723},
	url = {https://www.nature.com/articles/s41467-023-36159-y},
	doi = {10.1038/s41467-023-36159-y},
	language = {en},
	number = {1},
	urldate = {2026-01-16},
	journal = {Nature Communications},
	publisher = {Nature Publishing Group},
	author = {Jerbi, Sofiene and Fiderer, Lukas J. and Poulsen Nautrup, Hendrik and Kübler, Jonas M. and Briegel, Hans J. and Dunjko, Vedran},
	month = jan,
	year = {2023},
	pages = {517},
}

@article{havlicek_supervised_2019,
	title = {Supervised learning with quantum-enhanced feature spaces},
	volume = {567},
	copyright = {2019 The Author(s), under exclusive licence to Springer Nature Limited},
	issn = {1476-4687},
	url = {https://www.nature.com/articles/s41586-019-0980-2},
	doi = {10.1038/s41586-019-0980-2},
	language = {en},
	number = {7747},
	urldate = {2026-01-16},
	journal = {Nature},
	publisher = {Nature Publishing Group},
	author = {Havlíček, Vojtěch and Córcoles, Antonio D. and Temme, Kristan and Harrow, Aram W. and Kandala, Abhinav and Chow, Jerry M. and Gambetta, Jay M.},
	month = mar,
	year = {2019},
	pages = {209--212},
}

@article{rath_quantum_2024,
	title = {Quantum data encoding: a comparative analysis of classical-to-quantum mapping techniques and their impact on machine learning accuracy},
	volume = {11},
	issn = {2196-0763},
	shorttitle = {Quantum data encoding},
	url = {https://doi.org/10.1140/epjqt/s40507-024-00285-3},
	doi = {10.1140/epjqt/s40507-024-00285-3},
	language = {en},
	number = {1},
	urldate = {2026-01-16},
	journal = {EPJ Quantum Technology},
	author = {Rath, Minati and Date, Hema},
	month = oct,
	year = {2024},
	pages = {72},
}

@article{thanasilp_exponential_2024,
	title = {Exponential concentration in quantum kernel methods},
	volume = {15},
	copyright = {2024 The Author(s)},
	issn = {2041-1723},
	url = {https://www.nature.com/articles/s41467-024-49287-w},
	doi = {10.1038/s41467-024-49287-w},
	language = {en},
	number = {1},
	urldate = {2026-01-16},
	journal = {Nature Communications},
	publisher = {Nature Publishing Group},
	author = {Thanasilp, Supanut and Wang, Samson and Cerezo, M. and Holmes, Zoë},
	month = jun,
	year = {2024},
	pages = {5200},
}

@article{glick_covariant_2024,
	title = {Covariant quantum kernels for data with group structure},
	volume = {20},
	copyright = {2024 The Author(s), under exclusive licence to Springer Nature Limited},
	issn = {1745-2481},
	url = {https://www.nature.com/articles/s41567-023-02340-9},
	doi = {10.1038/s41567-023-02340-9},
	language = {en},
	number = {3},
	urldate = {2026-01-16},
	journal = {Nature Physics},
	publisher = {Nature Publishing Group},
	author = {Glick, Jennifer R. and Gujarati, Tanvi P. and Córcoles, Antonio D. and Kim, Youngseok and Kandala, Abhinav and Gambetta, Jay M. and Temme, Kristan},
	month = mar,
	year = {2024},
	pages = {479--483},
}

@article{hubregtsen_training_2022,
	title = {Training quantum embedding kernels on near-term quantum computers},
	volume = {106},
	url = {https://link.aps.org/doi/10.1103/PhysRevA.106.042431},
	doi = {10.1103/PhysRevA.106.042431},
	number = {4},
	urldate = {2026-01-16},
	journal = {Physical Review A},
	publisher = {American Physical Society},
	author = {Hubregtsen, Thomas and Wierichs, David and Gil-Fuster, Elies and Derks, Peter-Jan H. S. and Faehrmann, Paul K. and Meyer, Johannes Jakob},
	month = oct,
	year = {2022},
	pages = {042431},
}

@article{sahin_efficient_2024,
	title = {Efficient {Parameter} {Optimisation} for {Quantum} {Kernel} {Alignment}: {A} {Sub}-sampling {Approach} in {Variational} {Training}},
	volume = {8},
	shorttitle = {Efficient {Parameter} {Optimisation} for {Quantum} {Kernel} {Alignment}},
	url = {https://quantum-journal.org/papers/q-2024-10-18-1502/},
	doi = {10.22331/q-2024-10-18-1502},
	language = {en-GB},
	urldate = {2026-01-16},
	journal = {Quantum},
	publisher = {Verein zur Förderung des Open Access Publizierens in den Quantenwissenschaften},
	author = {Sahin, M. Emre and Symons, Benjamin C. B. and Pati, Pushpak and Minhas, Fayyaz and Millar, Declan and Gabrani, Maria and Mensa, Stefano and Robertus, Jan Lukas},
	month = oct,
	year = {2024},
	pages = {1502},
}

@article{henderson_quantum_2024,
	title = {Quantum {Kernel} {Machine} {Learning} {With} {Continuous} {Variables}},
	volume = {8},
	url = {https://quantum-journal.org/papers/q-2024-12-17-1570/},
	doi = {10.22331/q-2024-12-17-1570},
	language = {en-GB},
	urldate = {2026-01-16},
	journal = {Quantum},
	publisher = {Verein zur Förderung des Open Access Publizierens in den Quantenwissenschaften},
	author = {Henderson, Laura J. and Goel, Rishi and Shrapnel, Sally},
	month = dec,
	year = {2024},
	pages = {1570},
}

@article{park_variational_2023,
	title = {Variational quantum approximate support vector machine with inference transfer},
	volume = {13},
	copyright = {2023 The Author(s)},
	issn = {2045-2322},
	url = {https://www.nature.com/articles/s41598-023-29495-y},
	doi = {10.1038/s41598-023-29495-y},
	language = {en},
	number = {1},
	urldate = {2026-01-16},
	journal = {Scientific Reports},
	publisher = {Nature Publishing Group},
	author = {Park, Siheon and Park, Daniel K. and Rhee, June-Koo Kevin},
	month = feb,
	year = {2023},
	pages = {3288},
}

@inproceedings{cristianini_kernel-target_2001,
	address = {Cambridge, MA, USA},
	series = {{NIPS}'01},
	title = {On kernel-target alignment},
	doi = {10.5555/2980539.2980588},
	urldate = {2026-01-16},
	booktitle = {Proceedings of the 15th {International} {Conference} on {Neural} {Information} {Processing} {Systems}: {Natural} and {Synthetic}},
	publisher = {MIT Press},
	author = {Cristianini, Nello and Shawe-Taylor, John and Elisseeff, Andre and Kandola, Jaz},
	month = jan,
	year = {2001},
	pages = {367--373},
}

@article{wu_application_2021,
	title = {Application of quantum machine learning using the quantum kernel algorithm on high energy physics analysis at the {LHC}},
	volume = {3},
	url = {https://link.aps.org/doi/10.1103/PhysRevResearch.3.033221},
	doi = {10.1103/PhysRevResearch.3.033221},
	number = {3},
	urldate = {2026-01-16},
	journal = {Physical Review Research},
	publisher = {American Physical Society},
	author = {Wu, Sau Lan and Sun, Shaojun and Guan, Wen and Zhou, Chen and Chan, Jay and Cheng, Chi Lung and Pham, Tuan and Qian, Yan and Wang, Alex Zeng and Zhang, Rui and Livny, Miron and Glick, Jennifer and Barkoutsos, Panagiotis Kl. and Woerner, Stefan and Tavernelli, Ivano and Carminati, Federico and Di Meglio, Alberto and Li, Andy C. Y. and Lykken, Joseph and Spentzouris, Panagiotis and Chen, Samuel Yen-Chi and Yoo, Shinjae and Wei, Tzu-Chieh},
	month = sep,
	year = {2021},
	pages = {033221},
}

@inproceedings{li_very_2006,
	address = {New York, NY, USA},
	series = {{KDD} '06},
	title = {Very sparse random projections},
	isbn = {978-1-59593-339-3},
	url = {https://dl.acm.org/doi/10.1145/1150402.1150436},
	doi = {10.1145/1150402.1150436},
	urldate = {2026-01-16},
	booktitle = {Proceedings of the 12th {ACM} {SIGKDD} international conference on {Knowledge} discovery and data mining},
	publisher = {Association for Computing Machinery},
	author = {Li, Ping and Hastie, Trevor J. and Church, Kenneth W.},
	month = aug,
	year = {2006},
	pages = {287--296},
}

@article{achlioptas_database-friendly_2003,
	series = {Special {Issue} on {PODS} 2001},
	title = {Database-friendly random projections: {Johnson}-{Lindenstrauss} with binary coins},
	volume = {66},
	issn = {0022-0000},
	shorttitle = {Database-friendly random projections},
	url = {https://www.sciencedirect.com/science/article/pii/S0022000003000254},
	doi = {10.1016/S0022-0000(03)00025-4},
	number = {4},
	urldate = {2026-01-16},
	journal = {Journal of Computer and System Sciences},
	author = {Achlioptas, Dimitris},
	month = jun,
	year = {2003},
	pages = {671--687},
}

@article{PhysRevA.103.032430,
  title = {Effect of data encoding on the expressive power of variational quantum-machine-learning models},
  author = {Schuld, Maria and Sweke, Ryan and Meyer, Johannes Jakob},
  journal = {Phys. Rev. A},
  volume = {103},
  issue = {3},
  pages = {032430},
  numpages = {12},
  year = {2021},
  month = {Mar},
  publisher = {American Physical Society},
  doi = {10.1103/PhysRevA.103.032430},
  url = {https://link.aps.org/doi/10.1103/PhysRevA.103.032430}
}

@inproceedings{kuebler_inductive_bias_quantum_kernels_2021,
author = {K\"{u}bler, Jonas M. and Buchholz, Simon and Sch\"{o}lkopf, Bernhard},
title = {The inductive bias of quantum kernels},
year = {2021},
isbn = {9781713845393},
publisher = {Curran Associates Inc.},
address = {Red Hook, NY, USA},
booktitle = {Proceedings of the 35th International Conference on Neural Information Processing Systems},
articleno = {969},
numpages = {13},
series = {NIPS '21}
}

@inproceedings{yi2025surveyft,
  title     = {A Survey on Deep Learning based Time Series Analysis with Frequency Transformation},
  author    = {Yi, Kun and Zhang, Qi and Fan, Wei and Cao, Longbing and Wang, Shoujin and He, Hui and Long, Guodong and Hu, Liang and Wen, Qingsong and Xiong, Hui},
  booktitle = {Proceedings of the 31st ACM SIGKDD Conference on Knowledge Discovery and Data Mining (KDD '25)},
  year      = {2025},
  doi       = {10.1145/3711896.3736571}
}

@article{fulcher2014hctsa,
  title   = {Highly comparative feature-based time-series classification},
  author  = {Fulcher, Ben D. and Jones, Nick S.},
  journal = {IEEE Transactions on Knowledge and Data Engineering},
  year    = {2014},
  howpublished = {arXiv:1401.3531}
}

@article{mckay_efficient_z_2017,
  title   = {Efficient Z gates for quantum computing},
  author  = {McKay, David C. and Wood, Christopher J. and Sheldon, Sarah and Chow, Jerry M. and Gambetta, Jay M.},
  journal = {Physical Review A},
  year    = {2017},
  volume  = {96},
  number  = {2},
  pages   = {022330},
  doi     = {10.1103/PhysRevA.96.022330}
}

@article{kjaergaard_superconducting_2020,
  title   = {Superconducting Qubits: Current State of Play},
  author  = {Kjaergaard, Morten and Schwartz, Mollie E. and Braum{\"u}ller, Jochen and Krantz, Philip and Wang, Joel I.-J. and Gustavsson, Simon and Oliver, William D.},
  journal = {Annual Review of Condensed Matter Physics},
  year    = {2020},
  volume  = {11},
  pages   = {369--395},
  doi     = {10.1146/annurev-conmatphys-031119-050605}
}

@article{vezvaee_virtualz_2025,
  title   = {Virtual-Z Gates and Symmetric Gate Compilation},
  author  = {Vezvaee, Arian and Tripathi, Vinay and Kowsari, Daria and Levenson-Falk, Eli and Lidar, Daniel A.},
  journal = {PRX Quantum},
  year    = {2025},
  volume  = {6},
  number  = {2},
  pages   = {020348},
  doi     = {10.1103/PRXQuantum.6.020348}
}

@article{welch_diagonal_unitaries_2014,
  title   = {Efficient quantum circuits for diagonal unitaries without ancillas},
  author  = {Welch, Jonathan and Greenbaum, Daniel and Mostame, Sarah and Aspuru-Guzik, Al{\'a}n},
  journal = {New Journal of Physics},
  year    = {2014},
  volume  = {16},
  pages   = {033040},
  doi     = {10.1088/1367-2630/16/3/033040}
}

@article{agliardi_mitigating_covariant_kernels_2025,
  title   = {Mitigating exponential concentration in covariant quantum kernels for subspace and real-world data},
  author  = {Agliardi, Gabriele and Cortiana, Giorgio and Dekusar, Anton and Ghosh, Kumar and Mohseni, Naeimeh and O’Meara, Corey and Valls, Víctor and Yogaraj, Kavitha and Zhuk, Sergiy},
  journal = {npj Quantum Information},
  year    = {2025},
  volume  = {12},
  pages   = {12},
  doi     = {10.1038/s41534-025-01154-2}
}

@misc{zhou2022fedformer,
  title        = {FEDformer: Frequency Enhanced Decomposed Transformer for Long-term Series Forecasting},
  author       = {Zhou, Tian and Ma, Ziqing and Wen, Qingsong and Wang, Xue and Sun, Liang and Jin, Rong},
  year         = {2022},
  howpublished = {arXiv:2201.12740},
  doi          = {10.48550/arXiv.2201.12740}
}

@misc{leethorp2021fnet,
  title        = {FNet: Mixing Tokens with Fourier Transforms},
  author       = {Lee-Thorp, James and Ainslie, Joshua and Eckstein, Ilya and Onta{\~n}{\'o}n, Santiago},
  year         = {2021},
  howpublished = {arXiv:2105.03824},
  doi          = {10.48550/arXiv.2105.03824}
}

@article{bremner_commuting_2010,
  title   = {Commuting quantum computations: post hoc distributed sampling and classical simulations},
  author  = {Bremner, Michael J. and Jozsa, Richard and Shepherd, Daniel J.},
  journal = {Proceedings of the Royal Society A},
  year    = {2010},
  volume  = {467},
  number  = {2126},
  pages   = {459--472},
  doi     = {10.1098/rspa.2010.0203}
}

@article{cooley_tukey_fft_1965,
  author       = {Cooley, J. W. and Tukey, John W.},
  title        = {An Algorithm for the Machine Calculation of Complex Fourier Series},
  journal      = {Mathematics of Computation},
  volume       = {19},
  number       = {90},
  pages        = {297--301},
  year         = {1965},
  doi          = {10.1090/S0025-5718-1965-0178586-1},
  note         = {Introduces the FFT algorithm reducing DFT complexity from $O(N^2)$ to $O(N\log N)$ via divide-and-conquer recursion} 
}

@inproceedings{bingham_random_projection_2001,
  author       = {Bingham, Ella and Mannila, Heikki},
  title        = {Random Projection in Dimensionality Reduction: Applications to Image and Text Data},
  booktitle    = {Proceedings of the Seventh ACM SIGKDD International Conference on Knowledge Discovery and Data Mining (KDD ’01)},
  year         = {2001},
  pages        = {245--250},
  publisher    = {ACM},
  note         = {Shows that random projections are computationally simpler than PCA and effective for dimensionality reduction with low distortion} 
}

@article{yang_how_to_reduce_2020,
  author       = {Yang, Fan and Liu, Sifan and Dobriban, Edgar and Woodruff, David P.},
  title        = {How to reduce dimension with PCA and random projections?},
  journal      = {arXiv preprint},
  volume       = {abs/2005.00511},
  year         = {2020},
  note         = {Reviews differences between PCA (data-aware) and random projections (data-oblivious), noting speed advantages of random projections for dimensionality reduction} 
}

@InProceedings{10.1007/978-3-032-08381-4_14,
author="Herrero-G{\'o}mez, Pablo
and Pujol-L{\'o}pez, Francisco
and Mart{\'i}nez Garc{\'i}a, Ana
and Mora Mora, Higinio",
editor="Le Thi, Hoai An
and Pham Dinh, Tao
and Le, Hoai Minh",
title="Quantum Computing Models for Financial Analysis",
booktitle="Modelling, Computation and Optimization in Information Systems and Management Sciences",
year="2026",
publisher="Springer Nature Switzerland",
address="Cham",
pages="183--194",
isbn="978-3-032-08381-4"
}

@InProceedings{10.1007/978-3-032-07992-3_20,
author="Alemany Manzanaro, Marcos
and Mora, Higinio",
editor="Arai, Kohei",
title="Assessing the Viability of Quantum SAT Solvers: A Comparative Study Using Grover's Algorithm and Quantum Hardware",
booktitle="Proceedings of the Future Technologies Conference (FTC) 2025, Volume 4",
year="2026",
publisher="Springer Nature Switzerland",
address="Cham",
pages="310--325",
isbn="978-3-032-07992-3"
}

\end{document}